\documentclass[jair,twoside,11pt,english,reqno,theapa]{article}
\input epsf %% at top of file

\usepackage{jair}
\usepackage{theapa}

\jairheading{27}{2006}{25-53}{10/05}{9/06}
\ShortHeadings{Generative Prior Knowledge for Discriminative Classification}
{Epshteyn \& DeJong}
\firstpageno{25}

\usepackage{graphicx}
\usepackage{amssymb}

\usepackage{amsthm}
\usepackage{amsmath}

\makeatletter
 \theoremstyle{plain}    
 \newtheorem{thm}{Theorem}[section]
 \newtheorem*{thm*}{Theorem}
 \newtheorem*{thma}{Theorem \ref{thm-gen-bound}}
\newtheorem*{thmb}{Theorem \ref{thm-iter-conv}}

 \numberwithin{equation}{section} %% Comment out for sequentially-numbered
 \numberwithin{figure}{section} %% Comment out for sequentially-numbered
 \theoremstyle{plain}
 \theoremstyle{plain}    
 \newtheorem{prop}[thm]{Proposition} %%Delete [thm] to re-start numbering
 \newtheorem{lem}[thm]{Lemma} %%Delete [thm] to re-start numbering
 \theoremstyle{definition}
 \newtheorem{defn}[thm]{Definition}
 \theoremstyle{plain}    
 \newtheorem{cor}[thm]{Corollary} %%Delete [thm] to re-start numbering

\newcommand{\such}{\mathop{\mathrm{s.t.}}} 
\newcommand{\solves}{\mathop{\mathrm{solves}}} 
\newcommand{\pd}[2]{\frac{\partial #1}{\partial #2}}

\makeatother
\begin{document}

\title{Generative Prior Knowledge for Discriminative Classification}
\author{\name Arkady Epshteyn \email aepshtey@uiuc.edu \\
\name Gerald DeJong \email dejong@uiuc.edu \\
\addr Department of Computer Science\\ University of Illinois at Urbana-Champaign\\ 201 N. Goodwin\\ Urbana, IL, 61801 USA}

\maketitle

\begin{abstract}
We present a novel framework for integrating prior knowledge into
discriminative classifiers. Our framework allows discriminative
classifiers such as Support Vector Machines (SVMs) to utilize prior knowledge
specified in the generative setting. The dual objective of fitting
the data and respecting prior knowledge is formulated as a bilevel
program, which is solved (approximately) via iterative application
of second-order cone programming. To test our approach, we consider
the problem of using WordNet (a semantic database of English language)
to improve low-sample classification accuracy of newsgroup categorization.
WordNet is viewed as an approximate, but readily available source
of background knowledge, and our framework is capable of utilizing
it in a flexible way.
\end{abstract}
\maketitle

\section{Introduction}

While SVM \cite{vapnik} 
classification accuracy on many classification
tasks is often competitive with that of human subjects, the number
of training examples required to achieve this accuracy is prohibitively
large for some domains. Intelligent user interfaces, for example,
must adopt to the behavior of an individual user after a limited amount
of interaction in order to be useful. Medical systems diagnosing rare
diseases have to generalize well after seeing very few examples. Any
natural language processing task that performs processing at the level
of n-grams or phrases (which is frequent in translation systems) cannot
expect to see the same sequence of words a sufficient number of times
even in large training corpora. Moreover, supervised classification
methods rely on manually labeled data, which can be expensive to
obtain. Thus, it is important to improve classification performance
on very small datasets. Most classifiers are not competitive with
humans in their ability to generalize after seeing very few examples.
Various techniques have been proposed to address this problem, such
as active learning 
\cite{svm-active-koller,svm-active}, 
hybrid generative-discriminative
classification 
\cite{ng:hybrid}, 
learning-to-learn by extracting
common information from related learning tasks 
\cite{thrun-learn2learn,baxter-learn2learn,single-example},
and using prior knowledge. 

In this work, we concentrate on improving small-sample classification
accuracy with prior knowledge. While prior knowledge has proven useful
for classification 
\cite{scholkopf,wu,fung,rotate-svm,qiang-ebl}, 
it is notoriously
hard to apply in practice because there is a mismatch between the
form of prior knowledge that can be employed by classification algorithms
(either prior probabilities or explicit constraints on the hypothesis
space of the classifier) and the domain theories articulated by human
experts. This is unfortunate because various ontologies and domain
theories are available in abundance, but considerable amount of manual
effort is required to incorporate existing prior knowledge into the native learning bias
of the chosen algorithm. What would it take to apply an existing
domain theory automatically to a classification task for which it
was not specifically designed? In this work, we take the first steps
towards answering this question.

In our experiments, such a domain theory is exemplified by WordNet,
a linguistic database of semantic connections among English words
\cite{Miller90}. 
We apply WordNet to a standard benchmark task of newsgroup categorization.
Conceptually, a generative model describes how the world works, while a discriminative model
is inextricably linked to a specific classification task. Thus, there is reason to believe that a 
generative interpretation of a domain theory would seem to be more natural and generalize better across different
classification tasks. In Section 2 we present empirical evidence that this is, indeed, the case with 
WordNet in the context of newsgroup classification.
For this reason, we interpret the domain theory in the generative setting. 
However, many successful learning algorithms (such as support vector machines) are discriminative. 
We present a framework which allows the use of generative prior in the
discriminative classification setting. 

Our algorithm assumes that the generative distribution of the data is given in the Bayesian
framework: $Prob(data|model)$ and the prior $Prob'(model)$ are known.  However, instead of 
performing Bayesian model averaging, we assume that a single model $M^*$ has been selected a-priori, and the 
observed data is a manifestation of that model (i.e., it is drawn according to $Prob(data|M^*)$).
The goal of the learning algorithm is to estimate $M^*$.  This estimation is performed as a two-player 
sequential game of full information. The bottom (generative) player chooses the Bayes-optimal
discriminator function $f(M)$ for the probability distribution $Prob(data|model=M)$ (\emph{without taking the training data into account}) 
given the model $M$. The model $M$ is chosen by the top (discriminative) player
 in such a way that its prior probability of occurring, given by $Prob'(M)$, is high, \emph{and} it forces the 
bottom player to minimize the training-set error of its Bayes-optimal discriminator $f(M)$. 
This estimation procedure gives rise to a bilevel program. We show that, while the problem is 
known to be NP-hard, its approximation can be solved efficiently by iterative application
of second-order cone programming. 

The only remaining issue is how to construct the generative prior $Prob'(model)$ automatically 
from the domain theory.  We describe how to solve this problem in Section 2, where we also argue that  
the generative setting is appropriate for capturing expert knowledge, employing WordNet as an 
illustrative example. In Section 3, we give the necessary preliminary
information and important known facts and definitions. Our framework
for incorporating generative prior into discriminative classification
is described in detail in Section 4.
We demonstrate the efficacy of our approach experimentally by presenting
the results of using WordNet for newsgroup classification in Section
5. A theoretical explanation of the improved generalization ability
of our discriminative classifier constrained by generative prior knowledge
appears in Section 6.  Section 7 describes related work. Section 8 concludes the paper and outlines directions for future research.

\section{Generative vs. Discriminative Interpretation of Domain Knowledge}

WordNet can be viewed as a network, with nodes representing words
and links representing relationships between two words (such as synonyms,
hypernyms (is-a), meronyms (part-of), etc.). An important property of
WordNet is that of semantic distance - the length (in links) of the
shortest path between any two words. Semantic distance approximately
captures the degree of semantic relatedness of two words. We set up an
experiment to evaluate the usefulness of WordNet for the task of 
newsgroup categorization.  Each posting was represented by a bag-of-words,
with each binary feature representing the presence of the corresponding word.
The evaluation was done on pairwise classification tasks in the following two settings:
\begin{enumerate}
\item The generative framework assumes that each posting $x=[x^{1},..,x^{n}]$ is generated by a distinct probability distribution for each newsgroup.  The simplest version of a Linear Discriminan Analysis (LDA) classifier posits that $x|(y=-1) \sim N(\mu_{1},I)$ and $x|(y=1) \sim N(\mu_{2},I)$ for posting $x$ given label $y\in \{-1,1\}$, where $I\in\mathbf{\mathbb{{R}}}^{(n\times n)}$ is the identity matrix. 
Classification is done by assigning the most probable label to $x$: $y(x)=1 \Leftrightarrow Prob(x|1)>Prob(x|-1)$.  It is well-known \cite<e.g. see>{duda} that this decision rule is equivalent to the one given by the hyperplane $(\mu_{2}-\mu_{1})^{T}x-\frac{1}{2}(\mu_{2}^{T}\mu_{2}-\mu_{1}^{T}\mu_{1})>0$.  The means $\widehat{\mu_{i}}=[\widehat{\mu_{i}^{1}},..,\widehat{\mu_{i}^{n}}]$ are estimated via maximum likelihood from the training data $[x_{1}, y_{1}],..,[x_{m},y_{m}]$\footnote{The standard LDA classifier assumes that $x|(y=-1)\sim N(\mu_{1},\Sigma)$ and  $x|(y=-1)\sim N(\mu_{2},\Sigma)$ and estimates the covariance matrix $\Sigma$ as well as the means $\mu_{1},\mu_{2}$ from the training data.  In our experiments, we take $\Sigma=I$.}.
\item The discriminative SVM classifier sets the separating hyperplane to directly minimize the number of errors on the training data: \\$[\widehat{w}=[\widehat{w^{1}},..,\widehat{w^{n}}],\widehat{b}]=\arg\min_{w,b}\left\Vert w \right\Vert \such y_{i}(w^{T}x_{i}+b)\geq 1, i=1,..,m$.
\end{enumerate}

Our experiment was conducted in the learning-to-learn framework
\cite{thrun-learn2learn,baxter-learn2learn,single-example}.
In the first stage, each classifier was trained using training data from 
the \emph{training task} (e.g., for classifying postings into 
the newsgroups 'atheism' and 'guns').  In the second stage,
the classifier was generalized using WordNet's semantic information.
In the third stage, the generalized classifier was applied to a different,
\emph{test task} (e.g., for classifying postings for the newsgroups 'atheism' vs. 'mideast') \emph{without seeing any data from this new classification task}.  
The only way for a classifier to generalize in this
setting is to use the original sample to acquire information about
WordNet, and then exploit this information to help it label examples from
the test sample. In learning how to perform this task, the system also learns how to utilize the classification knowledge implicit in WordNet.

We now describe the second and third stages for the two classifiers in more detail:
\begin{enumerate}
\item It is intuitive to interpret information embedded in WordNet as follows:
if the title of the newsgroup is 'guns', then all the
words with the same semantic distance to 'gun' (e.g., 'artillery',
'shooter', and 'ordnance' with the distance of two) provide a similar degree of classification information.  To quantify this intuition, let $l_{i,train}=[l^{1}_{i,train},..,l^{j}_{i,train},..,l^{n}_{i,train}]$ be the vector of semantic distances in WordNet between each feature word $j$ and the label of each training task newsgroup $i\in \{1,2\}$.  Define $\chi_{i}(v) \triangleq \frac{\sum_{j:l_{i,train}^{j}=v}\widehat{\mu_{i}^{j}}}{|j:l_{i,train}^{j}=v|}, i=1,2$, where $|\cdot|$ denotes cardinality of a set. 
$\chi_{i}$ compresses information in $\widehat{\mu_{i}}$ based on the assumption that words equidistant from the newsgroup label are equally likely to appear in a posting from that newsgroup.  
To test the performance of this compressed classifier on a new task with semantic distances given by $l_{i,test}$, the generative distributions are reconstructed via $\mu_{i}^{j}:=\chi_{i}(l^{j}_{i,test})$.  Notice that if the classifier is trained and tested on the same task, applying the function $\chi_{i}$ is equivalent to averaging the components of the means of the generative distribution corresponding to the equivalence classes of words equidistant from the label.  If the classifier is tested on a different classification task, the reconstruction process reassigns the averages based on the semantic distances to the new labels.

\item It is less intuitive to interpret WordNet in a discriminative setting.  One possible interpretation is that coefficients $w^{j}$ of the separating hyperplane are governed by semantic distances to labels, as captured by the compression function $\chi'(v,u)\triangleq\frac{\sum_{j:l_{1,train}^{j}=v,l_{2,train}^{j}=u} \widehat{w^{j}}}{|j:l_{1,train}^{j}=v,l_{2,train}^{j}=u|}$ and reconstructed via $w^{j}:=\chi'(l^{j}_{1,test},l^{j}_{2,test})$.

\end{enumerate}

\begin{figure}
\begin{center}
\begin{minipage}[t]{0.33\textwidth}
1)Train: atheism vs. guns 

~~Test: atheism vs. mideast 
\includegraphics[%
  scale=0.35]{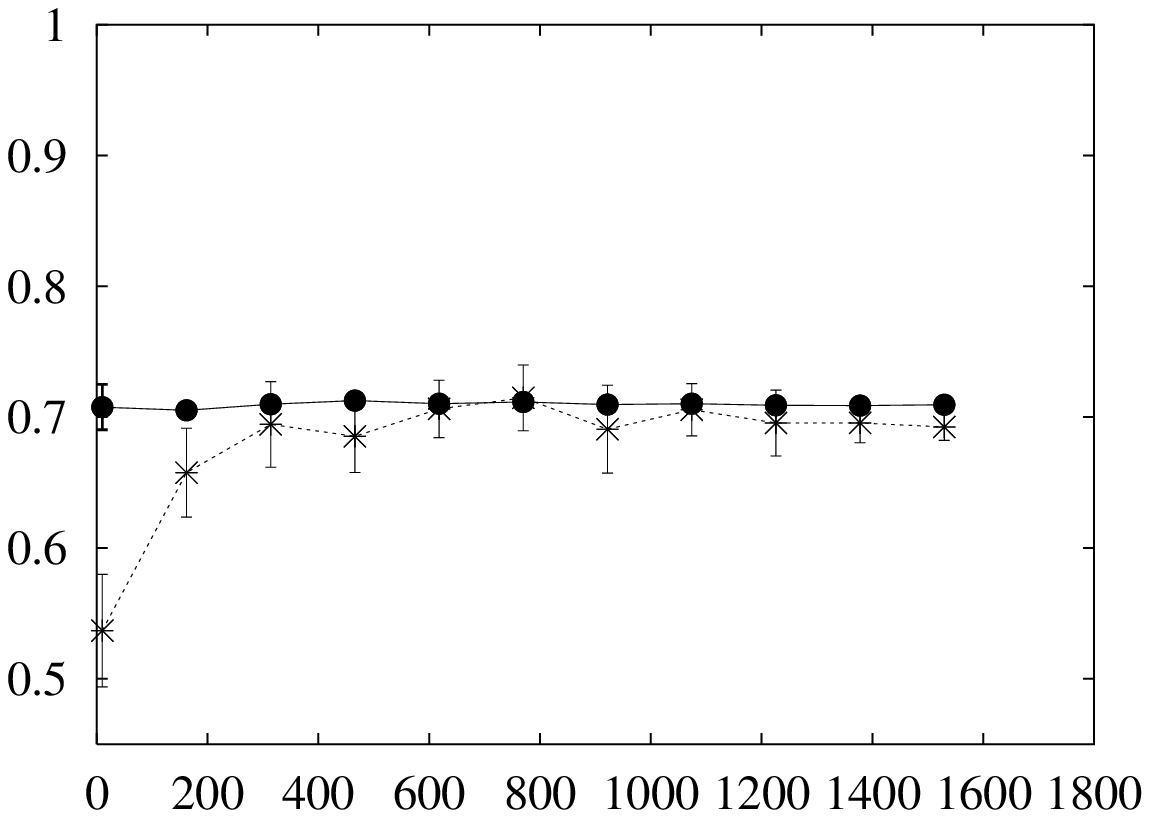}
\end{minipage}~~~\begin{minipage}[t]{0.33\textwidth}
2)Train: atheism vs. guns 

~~Test: guns vs. mideast
\includegraphics[%
  scale=0.35]{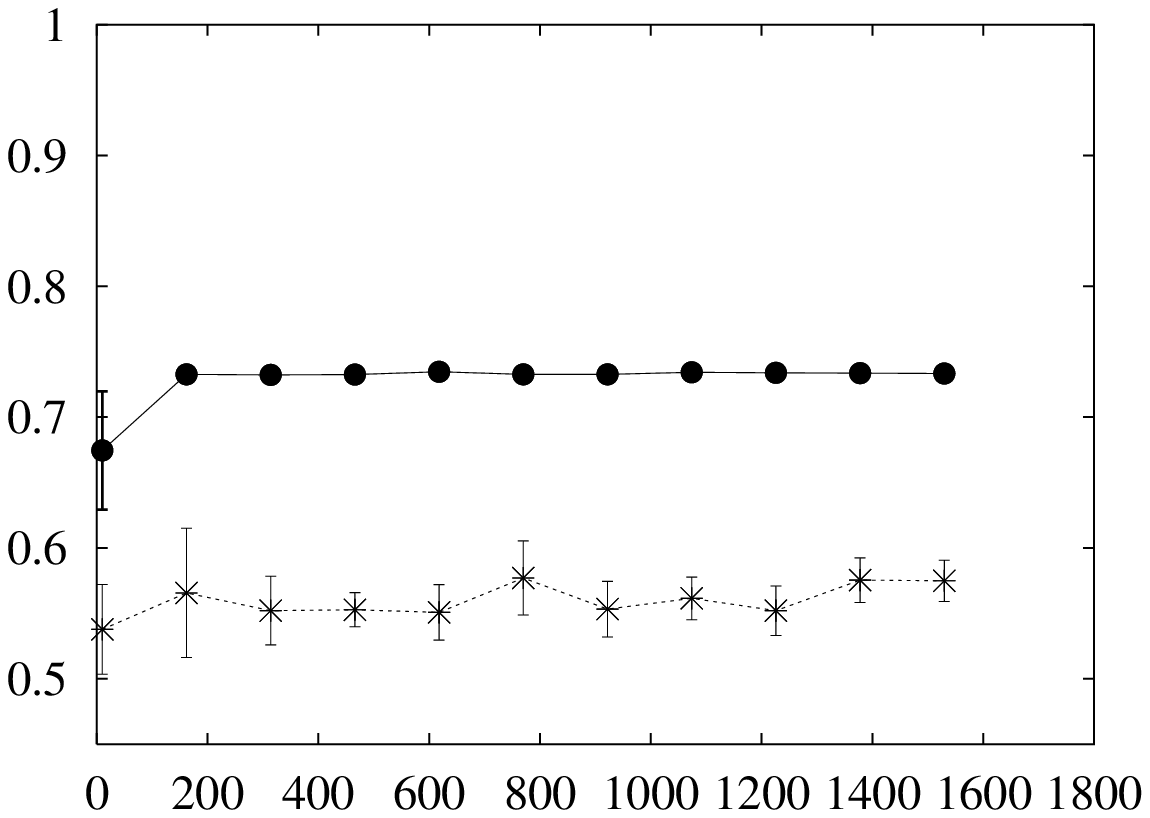}
\end{minipage}~~~\begin{minipage}[t]{0.33\textwidth}
3)Train: guns vs. mideast 

~~Test: atheism vs. mideast 
\includegraphics[%
  scale=0.35]{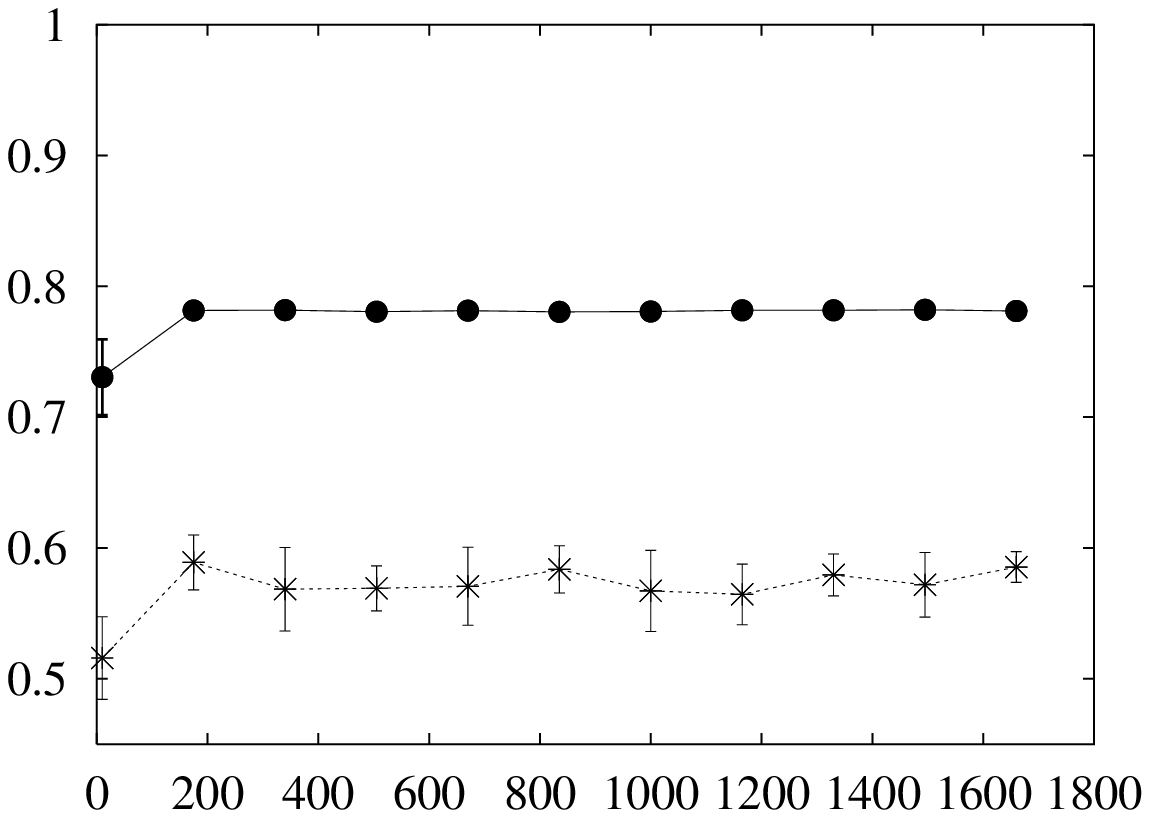}
\end{minipage}
\end{center}

\begin{center}
\includegraphics[%
  scale=0.35]{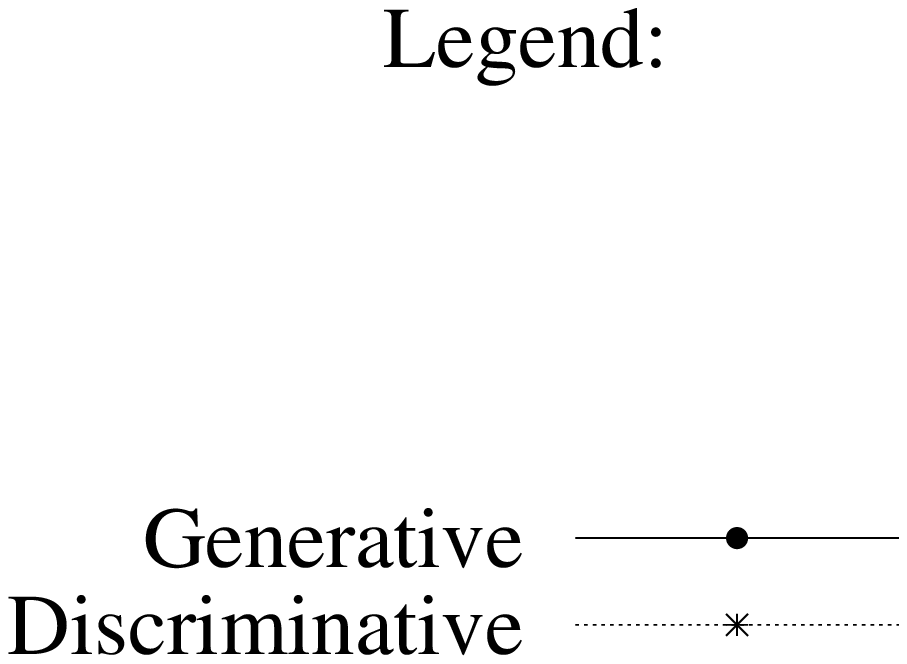}
\end{center}

\caption{Test set accuracy as a percentage versus the number of training points
for 3 different classification experiments. For each classification
task, a random test set is chosen from the full set of articles in
20 different ways. Error bars are based on 95\% confidence intervals.}
\label{cmp:discr-gen}
\end{figure}

Note that both the LDA generative classifier
and the SVM discriminative classifier have the same hypothesis space of separating hyperplanes.
The resulting test set classification accuracy for each classifier for a few classification tasks from the 20-newsgroup dataset \cite{20news}
is presented in Figure \ref{cmp:discr-gen}. The x-axis of each graph represents the size of the training task
sample, and the y-axis - the classifier's performance on the test
classification task. The generative classifier consistently outperforms
the discriminative classifier. It converges much faster, and on two
out of three tasks the discriminative classifier is not able to use
prior knowledge nearly as effectively as the generative classifier even
after seeing 90\% of all of the available training data. The generative
classifier is also more consistent in its performance - note that
its error bars are much smaller than those of the discriminative classifier.
%The fast convergence of the generative classifier can be explained
%by the fact that its equivalence classes (defined by the distance
%from the word to one of the labels) are larger than those of the discriminative
%classifier (defined by a pair of distances from the word to the two
%labels). As a direct result of this pooling, the statistical estimations are more reliable.
The results clearly show the potential of using background knowledge as a vehicle for sharing information 
between tasks. But the effective sharing is contingent on an appropriate task decomposition, here supplied by the tuned generative model.  

The evidence in Figure \ref{cmp:discr-gen} seemingly contradicts the conventional wisdom that discriminative training outperforms generative for sufficiently large training samples.  However, our experiment evaluates the two frameworks in the context of using an ontology to transfer information between learning tasks.  This was never done before. The experiment demonstrates that the interpretation of semantic distance in WordNet is more intuitive in the generative classification setting, probably because it better reflects the human intuitions behind WordNet. 

However, our goal
is not just to construct a classifier that performs well without seeing
\emph{any} examples of the test classification task. We also want a classifier
that improves its behavior as it sees new labeled data from the test
classification task. This presents us with a problem: one of the best-performing
classifiers \cite<and certainly the best on the text classification task
according to the study by>{joachims98text}
is SVM, a discriminative
classifier. Therefore, in the rest of this work, we focus on incorporating
generative prior knowledge into the discriminative classification
framework of support vector machines.

\section{Preliminaries}

It has been observed that constraints
on the probability measure of a half-space can be captured by second-order cone
constraints for Gaussian distributions \cite<see, e.g., the tutorial by>{lobo1}.  This allows for efficient processing
of such constraints within the framework of second-order cone programming (SOCP).  
We intend to model prior knowledge with elliptical distributions, a family of
probability distributions which generalizes Gaussians. In what follows, we
 give a brief overview of second-order cone programming and its relationship to constraints
imposed on the Gaussian probability distribution. We also note that it is possible to extend
the argument presented by Lobo et al. \citeyear{lobo1} to elliptical distributions.

Second-order cone program is a mathematical program of the form:
\begin{align}
\min_{x} & \text{ }v^{T}x\\
\such & \left\Vert A_{i}x+b_{i}\right\Vert \leq c_{i}^{T}x+d_{i},~i=1,...,N 
\end{align}
where $x\in\mathbb{{R}}{}^{n}$ is the optimization variable and $v\in\mathbf{\mathbb{{R}}}^{n}$,
$A_{i}\in\mathbb{{R}}{}^{(k_{i}\textrm{x}n)}$, $b_{i}\in\mathbb{{R}}{}^{k_{i}}$,
$c_{i}\in\mathbb{{R}}{}^{n}$ , $d_{i}\in\mathbb{{R}}$ are problem
parameters ($\left\Vert \cdot \right\Vert $ represents the usual
$L_{2}$-norm in this paper). SOCPs can be solved efficiently with
interior-point methods, as described by Lobo et al.
\citeyear{lobo1} in a tutorial which
contains an excellent overview of the theory and applications of SOCP.

We use the elliptical distribution to model distribution of the data
a-priori. Elliptical distributions are distributions with ellipsoidally-shaped
equiprobable contours. The density function of the $n$-variate elliptical
distribution has the form $f_{\mu,\Sigma,g}(x)=c(\text{det }\Sigma)^{-1}g((x-\mu)^{T}\Sigma^{-1}(x-\mu))$,
where $x\in\mathbb{{R}}{}^{n}$ is the random variable, $\mu\in\mathbb{{R}}{}^{n}$
is the location parameter, $\Sigma\in\mathbb{{R}}{}^{(n\textrm{x}n)}$
is a positive definite $(n\times n)$-matrix representing the
scale parameter, function $g(\cdot)$ is the density generator,
and $c$ is the normalizing constant. We will use the notation $X\sim E(\mu,\Sigma,g)$
to denote that the random variable $X$ has an elliptical distribution
with parameters $\mu,\Sigma,g$. Choosing appropriate density generator
functions $g$, the Gaussian distribution, the Student-t distribution,
the Cauchy distribution, the Laplace distribution, and the logistic
distribution can be seen as special cases of the elliptical distribution.
Using an elliptical distribution relaxes the restrictive assumptions
the user has to make when imposing a Gaussian prior, while keeping
many desirable properties of Gaussians, such as: 

\begin{enumerate}
\item If $X\sim E(\mu,\Sigma,g)$, $A\in\mathbb{{R}}{}^{(k\times n)}$,
and $B\in\mathbb{{R}}{}^{k}$, then $AX+B\sim E(A\mu+B,A\Sigma A^{T},g)$ 
\item If $X\sim E(\mu,\Sigma,g)$, then $E(X)=\mu$. 
\item If $X\sim E(\mu,\Sigma,g)$, then $Var(X)=\alpha_{g}\Sigma$, where
$\alpha_{g}$ is a constant that depends on the density generator
$g$. 
\end{enumerate}
The following proposition shows that for elliptical distributions,
the constraint $P(w^{T}x+b\geq0)\leq\eta$ (i.e., the probability
that $X$ takes values in the half-space $\{ w^{T}x+b\geq0\}$ is less
than $\eta$) is equivalent to a second-order cone constraint for
$\eta\leq\frac{1}{2}$:

\begin{prop}
\label{prop_socp1}
If $X\sim E(\mu,\Sigma,g),$ $Prob(w^{T}x+b\geq0)\leq\eta\leq\frac{1}{2}$
is equivalent to $-(w^{T}\mu+b)/\beta_{g,\eta}\geq\left\Vert \Sigma^{1/2}w\right\Vert $,
where $\beta_{g,\eta}$ is a constant which only depends on $g$ and
$\eta$.
\begin{proof}
The proof is identical to the one given by Lobo \citeyear{lobo1} and Lanckriet et al. \citeyear{minimax1} for Gaussian distributions and is provided here for completeness:
\begin{equation}
\label{eqn1}
\textrm{Assume }Prob(w^{T}x+b\geq0)\leq\eta.
\end{equation}
Let $u=w^{T}x+b$. Let $\overline{u}$ 
denote the mean of $u$, and $\sigma$ denote its variance. Then the
constraint \ref{eqn1} can be written as 
\begin{equation}
\label{eqn2}
Prob(\frac{u-\overline{u}}{\sqrt{\sigma}}\geq-\frac{\overline{u}}{\sqrt{\sigma}})\leq\eta.
\end{equation}
By the properties of elliptical distributions, $\overline{u}=w^{T}\mu+b$,
$\sigma=\sqrt{\alpha_{g}}\left\Vert \Sigma^{1/2}w\right\Vert $, and
$\frac{u-\overline{u}}{\sqrt{\sigma}}\sim E(0,1,g)$. Thus, statement
\ref{eqn2} above can be expressed as $Prob_{X\sim E(0,1,g)}(X\geq-\frac{w^{T}\mu+b}{\sqrt{\alpha_{g}}\left\Vert \Sigma^{1/2}w\right\Vert })\leq\eta$,
which is equivalent to $-\frac{w^{T}\mu+b}{\sqrt{\alpha_{g}}\left\Vert \Sigma^{1/2}w\right\Vert }\geq\Phi^{-1}(\eta)$,
where $\Phi(z)=Prob_{X\sim E(0,1,g)}(X\geq z)$. The proposition
follows with $\beta_{g,\eta}=\sqrt{\alpha_{g}}\Phi^{-1}(\eta)$.
\end{proof}
\end{prop}
\begin{prop}
\label{prop_socp2}
For any monotonically decreasing $g$, $Prob_{X\sim E(\mu,\Sigma,g)}(x)\geq\delta$
is equivalent to $\left\Vert \Sigma^{-1/2}(x-\mu)\right\Vert \leq\varphi_{g,c,\Sigma}$,
where $\varphi_{g,c,\Sigma,\delta}=g^{-1}(\frac{\delta|\Sigma|}{c})$
is a constant which only depends on $g,c,\Sigma,\delta$.
\begin{proof}
Follows directly from the definition of $Prob_{X\sim E(\mu,\Sigma,g)}(x)$.
\end{proof}
\end{prop}
\section{Generative Prior via Bilevel Programming}
\label{ch-gen-discr}
We deal with the binary classification task: the classifier is a function
$f(x)$ which maps instances $x\in\mathbb{{R}}{}^{n}$ to labels $y\in\{-1,1\}$.
In the generative setting, the probability densities $Prob(x|y=-1;\mu_{1})$
and $Prob(x|y=1;\mu_{2})$ parameterized by $\mu=[\mu_{1},\mu_{2}]$
are provided (or estimated from the data), along with the prior probabilities
on class labels $\Pi(y=-1)$ and $\Pi(y=1)$, and the Bayes optimal
decision rule is given by the classifier 
\[f(x|\mu)=sign(Prob(x|y=-1;\mu_{1})\Pi(y=-1)-Prob(x|y=1;\mu_{2})\Pi(y=1)),\]
where $sign(x):=1$ if $x\geq0$ and $-1$ otherwise. In LDA, for instance,
the parameters $\mu_{1}$ and $\mu_{2}$ are the means of the two 
Gaussian distributions generating the data given each label.

Informally, our approach to incorporating prior knowledge is straightforward:
we assume a two-level hierarchical generative probability distribution
model. The low-level probability distribution of the data given the
label $Prob(x|y;\mu)$ is parameterized by $\mu$, which, in
turn, has a known probability distribution $Prob'(\mu)$. The goal
of the classifier is to estimate the values of the parameter vector
$\mu$ from the training set of labeled points $[x_{1},y_{1}]...[x_{m},y_{m}]$.
This estimation is performed as a two-player sequential game of full
information. The bottom (generative) player, given $\mu$, selects
the Bayes optimal decision rule $f(x|\mu)$. The top (discriminative)
player selects the value of $\mu$ which has a high probability
of occurring (according to $Prob'(\mu)$) \emph{and} which will
force the bottom player to select the decision rule which minimizes
the discriminative error on the training set. We now give a more formal
specification of this training problem and formulate it as a bilevel
program. Some of the assumptions are subsequently relaxed to enforce
both tractability and flexibility.

We use an elliptical distribution $E(\mu_{1},\Sigma_{1},g)$ to model
$X|y=-1$, and another elliptical distribution $E(\mu_{2},\Sigma_{2},g)$
to model $X|y=1$. If the parameters $\mu_{i},\Sigma_{i},i=1,2$
are known, the Bayes optimal decision rule \emph{restricted to the
class of linear classifiers}%
\footnote{A decision rule \emph{restricted to some class of classifiers} $H$
is optimal if its probability of error is no larger than that of any
other classifier \emph{in} $H$ 
\cite{Tong:restrictedbayes}.%
} of the form $f_{w,b}(x)=sign(w^{T}x+b)$ is given by $f(x)$ which
minimizes the probability of error among all linear discriminants:
$Prob(error)=Prob(w^{T}x+b\geq0|y=1)\Pi(y=1)+Prob(w^{T}x+b\leq0|y=-1)\Pi(y=-1)=\frac{1}{2}(Prob_{X\sim E(\mu_{1},\Sigma_{1},g)}(w^{T}x+b\geq0)+Prob_{X\sim E(\mu_{2},\Sigma_{2},g)}(w^{T}x+b\leq0))$,
assuming equal prior probabilities for both classes. We now model
the uncertainty in the means of the elliptical distributions $\mu_{i},i=1,2$
by imposing elliptical prior distributions on the locations of the
means: $\mu_{i}\sim E(t_{i},\Omega_{i},g),i=1,2$. In addition, to ensure the 
optimization problem is well-defined, we maximize the margin of the hyperplane
subject to the imposed generative probability constraints:
\begin{align}
\min_{\mu_{1},\mu_{2}} &\left\Vert w\right\Vert \\
\such & y_{i}(w^{T}x_{i}+b) \geq1,\, i=1,..,m \\
& Prob_{\mu_{i}\sim E(t_{i},\Omega_{i},g)}(\mu_{i}) \geq\delta,\, i=1,2 \\
& [w,b] \solves~\min_{w,b} [Prob_{X\sim E(\mu_{1},\Sigma_{1},g)}(w^{T}x+b\geq0)+Prob_{X\sim E(\mu_{2},\Sigma_{2},g)}(w^{T}x+b\leq0)] \label{eqn_opt2}
\end{align}

This is a bilevel mathematical program (i.e., an optimization problem in which the constraint region is implicitly defined by another optimization problem), 
which is strongly NP-hard
even when all the constraints and both objectives are linear 
\cite{hansen-bilevel}.
However, we show that it is possible to solve a reasonable approximation
of this problem efficiently with several iterations of second-order
cone programming. First, we relax the second-level minimization (\ref{eqn_opt2})
by breaking it up into two constraints: $Prob_{X\sim E(\mu_{1},\Sigma_{1},g)}(w^{T}x+b\geq0)\leq\eta$
and $Prob_{X\sim E(\mu_{2},\Sigma_{2},g)}(w^{T}x+b\leq0)\leq\eta$. Thus,
instead of looking for the Bayes optimal decision boundary, the algorithm
looks for a decision boundary with low probability of error, where
low error is quantified by the choice of $\eta$. 

Propositions \ref{prop_socp1} and \ref{prop_socp2} enable us to rewrite the optimization problem
resulting from this relaxation as follows :
\begin{align}
\min_{\mu_{1},\mu_{2},w,b} & \left\Vert w\right\Vert \label{eqn_opt}\\
\such & y_{i}(w^{T}x_{i}+b)\geq1,\, i=1,..,m \label{eqn1_constr1}\\
& Prob_{\mu_{i}\sim E(t_{i},\Omega_{i},g)}(\mu_{i})\geq\delta,\, i=1,2~\Leftrightarrow~\left\Vert \Omega_{i}^{-1/2}(\mu_{i}-t_{i})\right\Vert \leq\varphi,i=1,2 \label{eqn_means}\\
& Prob_{X\sim E(\mu_{1},\Sigma_{1},g)}(w^{T}x+b\geq0)\leq\eta~\Leftrightarrow~-\frac{w^{T}\mu_{1}+b}{\left\Vert \Sigma_{1}^{1/2}w\right\Vert }\geq\beta \label{eqn1_constr3}\\
& Prob_{X\sim E(\mu_{2},\Sigma_{2},g)}(w^{T}x+b\leq0)\leq\eta~\Leftrightarrow~\frac{w^{T}\mu_{2}+b}{\left\Vert \Sigma_{2}^{1/2}w\right\Vert }\geq\beta \label{eqn1_constr4}
\end{align}

Notice that the form of this program does not depend on the generator
function $g$ of the elliptical distribution - only constants $\beta$
and $\varphi$ depend on it.  $\varphi$ defines how far the system
is willing to deviate from the prior in its choice of a generative model, and $\beta$ bounds the tail probabilities of error (Type I and Type II) which the system will tolerate assuming its chosen generative model is correct.
These constants depend both on the specific generator $g$ and the amount of error the user is willing to tolerate.  In our experiments, we select the values
of these constants to optimize performance. Unless the user wants
to control the probability bounds through these constants, it is sufficient
to assume a-priori only that probability distributions (both prior
and hyper-prior) are elliptical, without making any further commitments.

Our algorithm solves the above problem by repeating the following two steps:

\begin{enumerate}
\item Fix the top-level optimization parameters $\mu_{1}$ and $\mu_{2}$.
This step combines the objectives of maximizing the margin of the
classifier on the training data and ensuring that the decision boundary
is (approximately) Bayes optimal with respect to the given generative
probability densities specified by the $\mu_{1},\mu_{2}$. 
\item Fix the bottom-level optimization parameters $w,b$. Expand the feasible
region of the program in step 1 as a function of $\mu_{1},\mu_{2}$.
This step fixes the decision boundary and pushes the means of the
generative distribution as far away from the boundary as the constraint
(\ref{eqn_means}) will allow. 
\end{enumerate}
The steps are repeated until convergence (in practice, convergence is detected when
the optimization parameters do not change appreciably from one iteration to the next).
Each step of the algorithm can be formulated as a second-order
cone program:

Step 1. Fix $\mu_{1}$ and $\mu_{2}$. Removing unnecessary constraints
from the mathematical program above and pushing the objective into
constraints, we get the following SOCP: 
\begin{align}
\min_{w,b} & \rho \label{eqn_opt_step1}\\
\such & \rho\geq\left\Vert w\right\Vert \\
& y_{i}(w^{T}x_{i}+b)\geq1,\, i=1,..,m \label{eqn_constr1}\\
& -\frac{w^{T}\mu_{1}+b}{\left\Vert \Sigma_{1}^{1/2}w\right\Vert }\geq\beta \label{eqn_constr2}\\
& \frac{w^{T}\mu_{2}+b}{\left\Vert \Sigma_{2}^{1/2}w\right\Vert }\geq\beta \label{eqn_constr3}
\end{align}

Step 2. Fix $w,b$ and expand the span of the feasible region, as
measured by $\frac{w^{T}\mu_{2}+b}{\left\Vert \Sigma_{2}^{1/2}w\right\Vert }-\frac{w^{T}\mu_{1}+b}{\left\Vert \Sigma_{1}^{1/2}w\right\Vert }$.
Removing unnecessary constraints, we get:
\begin{align}
\max_{\mu_{1},\mu_{2}} & \frac{w^{T}\mu_{2}+b}{\left\Vert \Sigma_{2}^{1/2}w\right\Vert }-\frac{w^{T}\mu_{1}+b}{\left\Vert \Sigma_{1}^{1/2}w\right\Vert }\label{eqn_opt_step2}\\
\such & \left\Vert \Omega_{i}^{-1/2}(\mu_{i}-t_{i})\right\Vert \leq\varphi,\, i=1,2 \label{eqn_constr4}
\end{align}

\begin{figure}
\begin{center}
1) ~~~~~~~~~~~~~~~~~~~~~~~~~~~~~~~~~~~~~~~~~~~~~2)

\includegraphics[%
  scale=0.3]{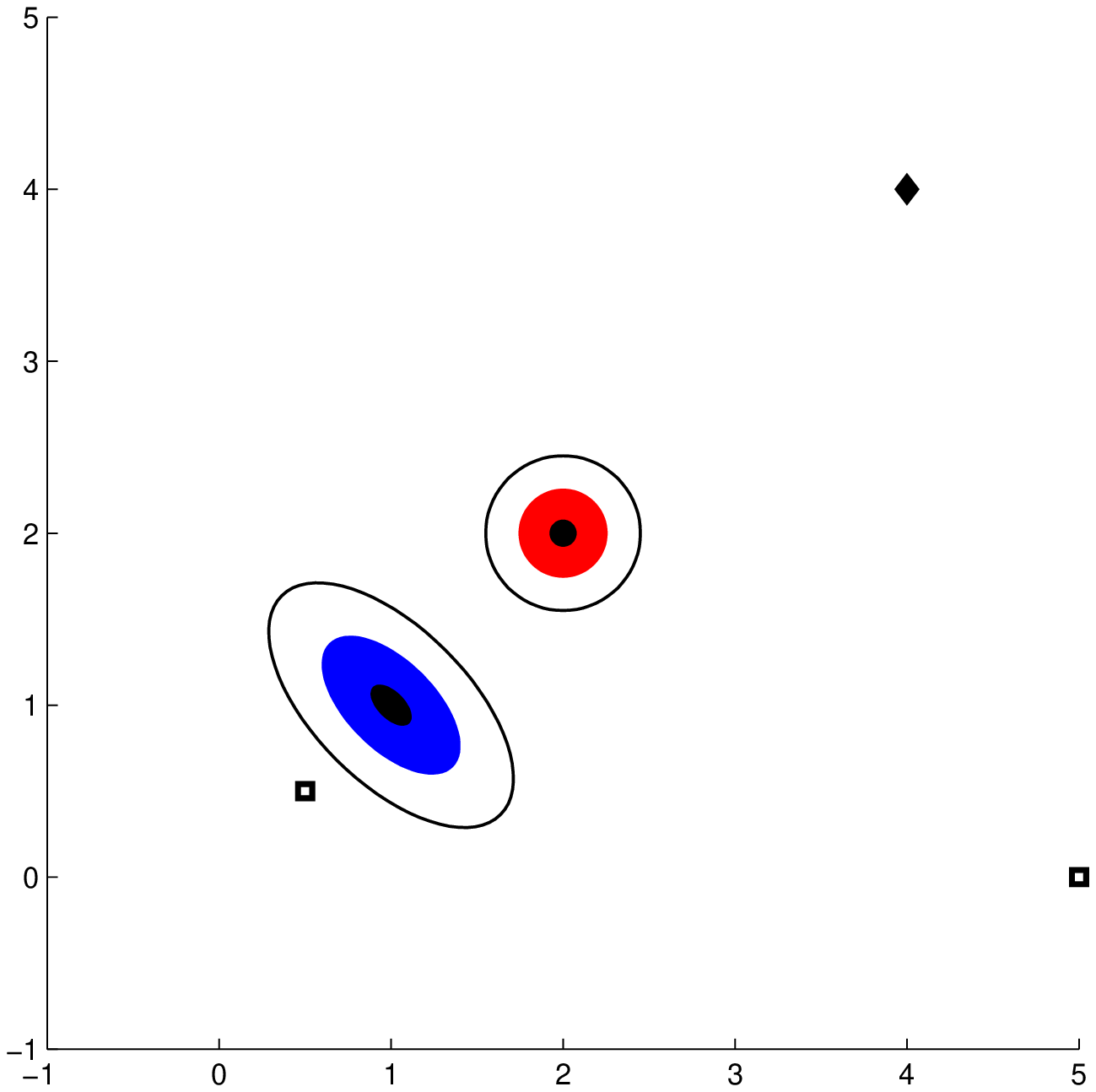}\includegraphics[%
  scale=0.3]{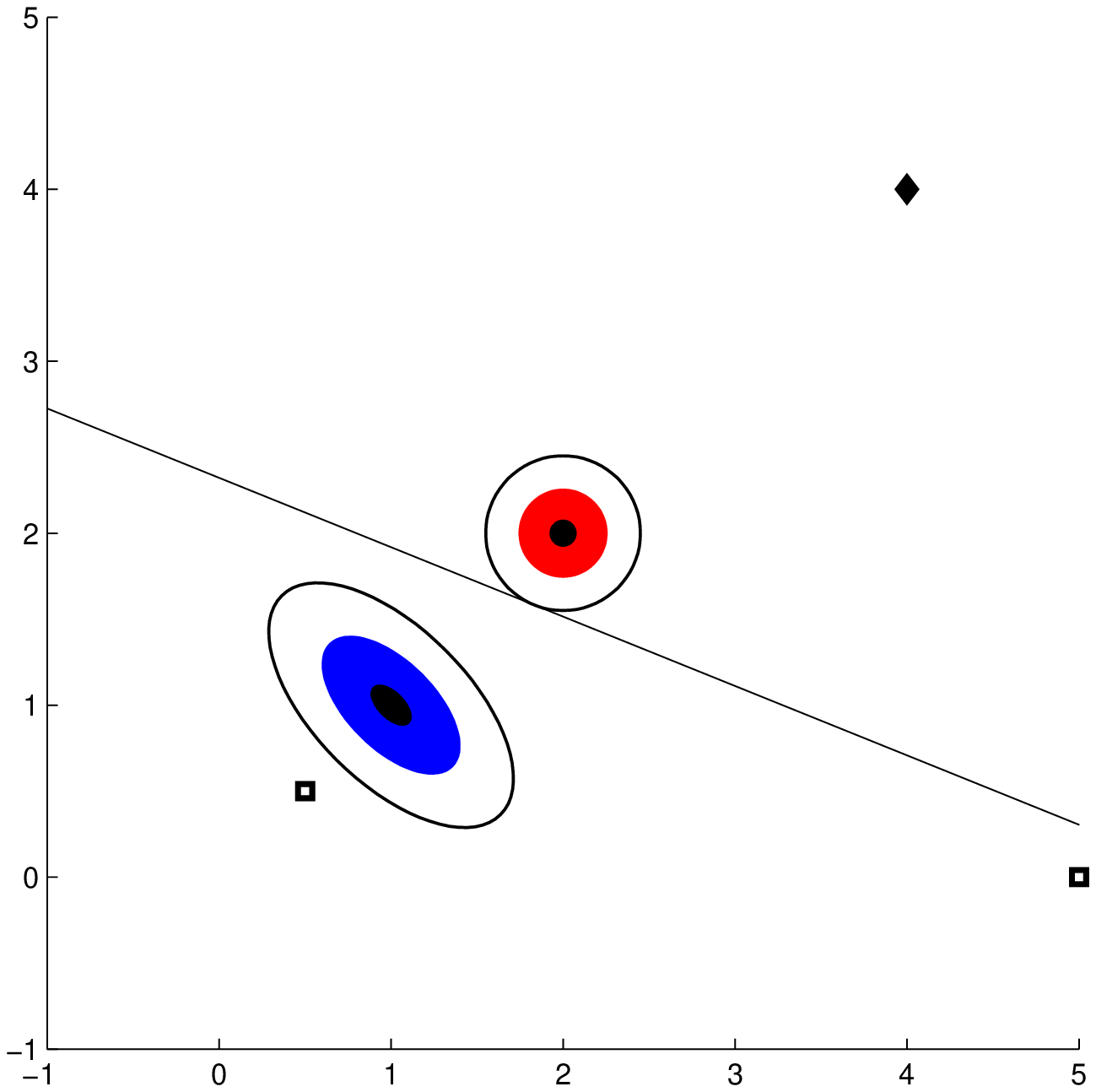}

3) ~~~~~~~~~~~~~~~~~~~~~~~~~~~~~~~~~~~~~~~~~~~~~4)

\includegraphics[%
  scale=0.3]{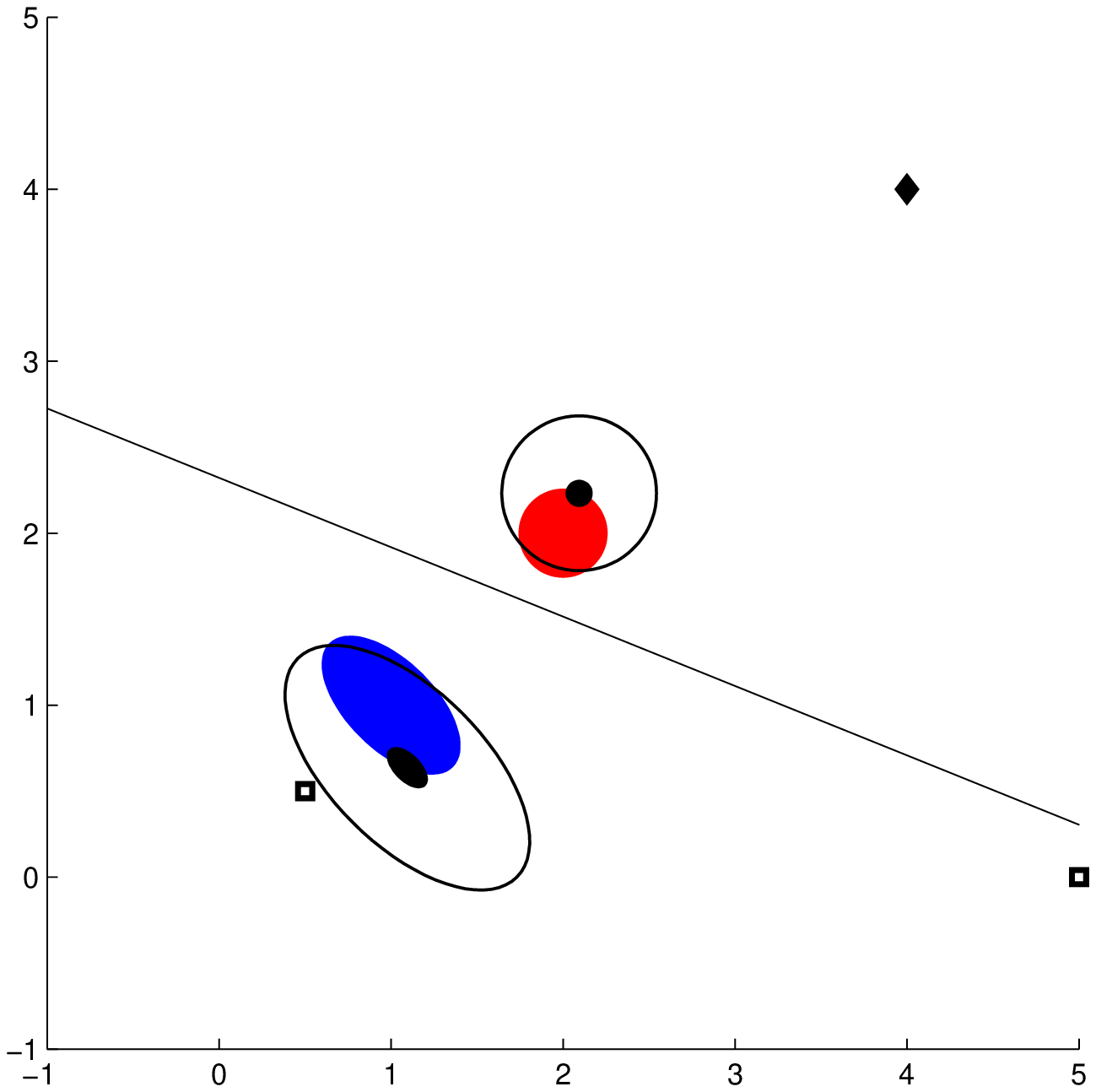}\includegraphics[%
  scale=0.3]{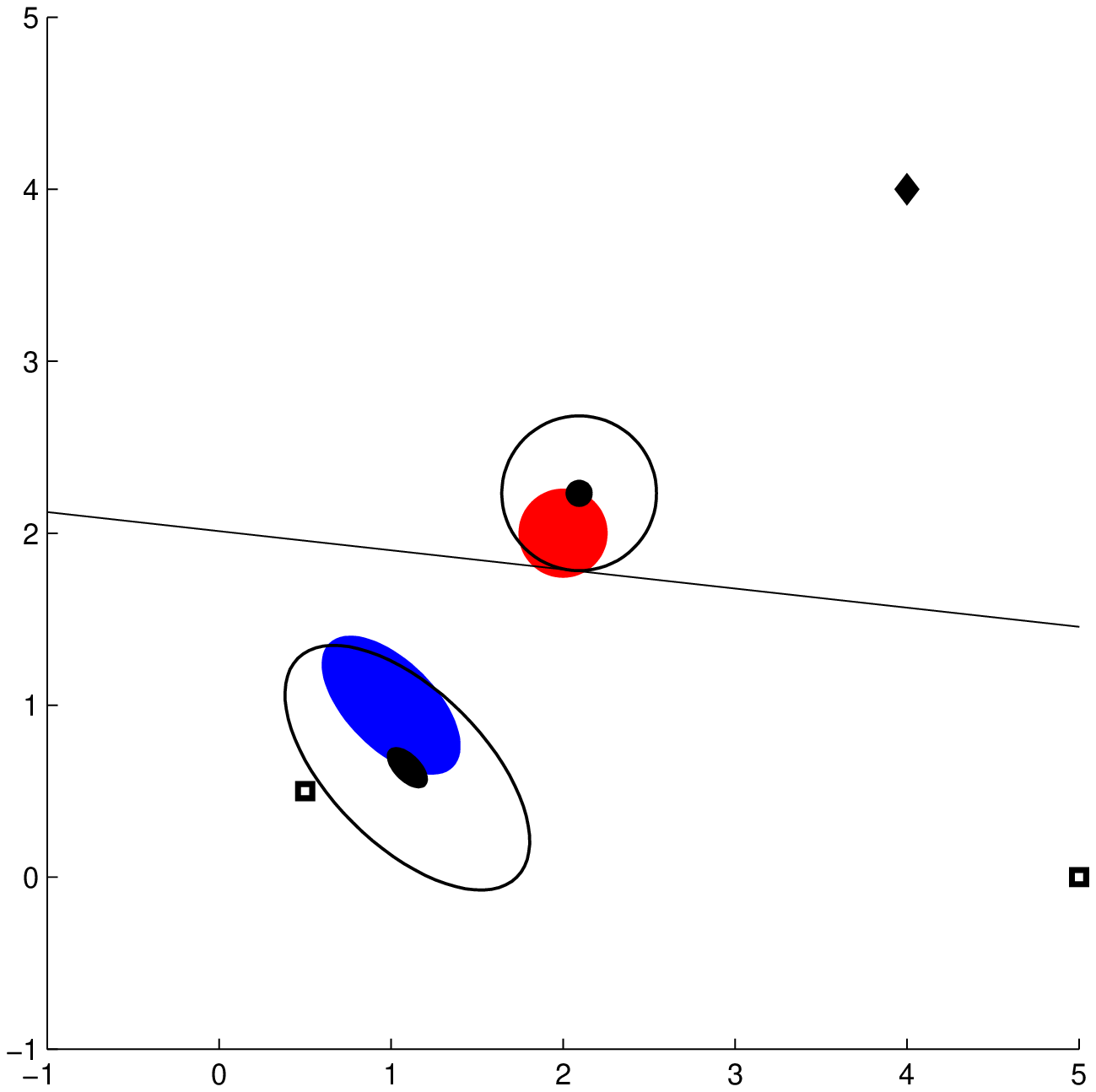}
\end{center}

\caption{Steps of the iterative (hard-margin) SOCP procedure:}
(The region where the hyperprior probability is larger than $\delta$ is shaded for each prior distribution. The covariance matrices are represented by equiprobable elliptical contours. In this example, the covariance matrices of the hyperprior and the prior distributions are multiples of each other. Data points from two different classes are represented by diamonds and squares.)

\begin{enumerate}
\item Data, prior, and hyperprior before the algorithm is executed. 
\item Hyperplane discriminator at the end of step 1, iteration 1
\item Priors at the end of step 2, iteration 1
\item Hyperplane discriminator at the end of step 2, iteration 2
\end{enumerate}
The algorithm converges at the end of step 2 for this problem 
(step 3 does not move the hyperplane).

\label{alg}
\end{figure}

The behavior of the algorithm is illustrated in Figure \ref{alg}.

The following theorems state that the algorithm converges.

\begin{thm}
\label{thm-obj-conv}
Suppose that the algorithm produces a sequence of iterates \\
$\left\{\mu_{1}^{(t)},\mu_{2}^{(t)},w^{(t)},b^{(t)}\right\}_{t=0}^{\infty}$, and the 
quality of each iterate is evaluated by its margin $\left\Vert w^{(t)} \right\Vert$.
This evaluation function converges.
\end{thm}
\begin{proof}
Let $\mu_{1}^{(t)},\mu_{2}^{(t)}$ be the values of the prior location parameters,
and $w_{1}^{(t)},b_{1}^{(t)}$ be the minimum error hyperplane the algorithm finds
at the end of the $t$-th step. At the end of the $(t+1)$-st step, $w_{1}^{(t+1)},b_{1}^{(t+1)}$
is still in the feasible region of the $t$-th step SOCP. This is true
because the function $f(\frac{(w^{(t)})^{T}\mu_{2}+b^{(t)}}{\left\Vert \Sigma_{2}^{1/2}w^{(t)}\right\Vert },-\frac{(w^{(t)})^{T}\mu_{1}+b^{(t)}}{\left\Vert \Sigma_{1}^{1/2}w^{(t)}\right\Vert })=\frac{(w^{(t)})^{T}\mu_{2}+b^{(t)}}{\left\Vert \Sigma_{2}^{1/2}w^{(t)}\right\Vert }-\frac{(w^{(t)})^{T}\mu_{1}+b^{(t)}}{\left\Vert \Sigma_{1}^{1/2}w^{(t)}\right\Vert }$
is monotonically increasing in each one of its arguments when the
other argument is fixed, and fixing $\mu_{1}$ (or $\mu_{2}$) fixes
exactly one argument. If the solution $\mu_{1}^{(t+1)},\mu_{2}^{(t+1)}$ at
the end of the $(t+1)$-st step were such that $\frac{(w^{(t)})^{T}\mu_{2}^{(t+1)}+b^{(t)}}{\left\Vert \Sigma_{2}^{1/2}w^{(t)}\right\Vert }<\beta$,
then $f$ could be increased by fixing $\mu_{1}^{(t+1)}$ and using the
value of $\mu_{2}^{(t)}$ from the beginning of the step which ensures that
$\frac{(w^{(t)})^{T}\mu_{2}^{(t)}+b^{(t)}}{\left\Vert \Sigma_{2}^{1/2}w^{(t)}\right\Vert }\geq\beta$,
which contradicts the observation that $f$ is maximized at the end
of the second step. The same contradiction is reached if $-\frac{(w^{(t)})^{T}\mu_{1}^{(t+1)}+b^{(t)}}{\left\Vert \Sigma_{1}^{1/2}w^{(t)}\right\Vert }<\beta$. 
Since the minimum error hyperplane from the previous iteration is
in the feasible region at the start of the next iteration, the objective
$\left\Vert w^{(t)}\right\Vert $ must decrease monotonically from one iteration
to the next. Since it is bounded below by zero, the algorithm converges. 
\end{proof}

In addition to the convergence of the objective function, the 
accumulation points of the sequence of iterates can be characterized by the following theorem:
\begin{thm}
\label{thm-iter-conv}
The accumulation points of the sequence $\left\{\mu_{1}^{(t)},\mu_{2}^{(t)},w^{(t)},b^{(t)}\right\}$ (i.e., limiting points of its convergent subsequences) have no feasible descent directions for the original optimization problem given by (\ref{eqn_opt})-(\ref{eqn1_constr4}).
\begin{proof}
See Appendix \ref{appendix-dual}.
\end{proof}
\end{thm}
If a point has no feasible descent directions, then any sufficiently small step along any directional vector will either increase the objective function, leave it unchanged, or take the algorithm outside of the feasible region.  The set of points with no feasible descent directions is a subset of the set of local minima.  Hence, convergence to such a point is a somewhat weaker result than convergence to a local minimum.

In practice, we observed rapid convergence usually within 2-4 iterations.

Finally, we may want to relax the strict assumptions of the correctness
of the prior/linear separability of the data by introducing slack
variables into the optimization problem above. This results in the
following program:
\begin{align}
\min_{\mu_{1},\mu_{2},w,b,\xi_{i},\zeta_{1},\zeta_{2},\nu_{1},\nu_{2}} & \left\Vert w\right\Vert +C_{1}\sum_{i=1}^{m}\xi_{i}+C_{2}(\zeta_{1}+\zeta_{2})+C_{3}(\nu_{1}+\nu_{2}) \\
\such & y_{i}(w^{T}x_{i}+b)\geq1-\xi_{i},\, i=1,..,m \\
& \left\Vert \Omega_{i}^{-1/2}(\mu_{i}-t_{i})\right\Vert \leq\varphi+\nu_{i},i=1,2 \\
& -\frac{w^{T}\mu_{1}+b}{\beta}\geq\left\Vert \Sigma_{1}^{1/2}w\right\Vert -\zeta_{1} \\
& \frac{w^{T}\mu_{2}+b}{\beta}\geq\left\Vert \Sigma_{2}^{1/2}w\right\Vert -\zeta_{2} \\
& \xi_{i}\geq0,~i=1,..,m \\
&\nu_{i}\geq0,~i=1,2 \\
& \zeta_{i}\geq0,~i=1,2
\end{align}

As before, this problem can be solved with the two-step iterative
SOCP procedure.  Imposing the generative prior with soft constraints ensures that, as the
amount of training data increases, the data overwhelms the prior and the algorithm converges to 
the maximum-margin separating hyperplane.

\section{Experiments}
\begin{figure}
\begin{center}
\includegraphics[%
  scale=0.75]{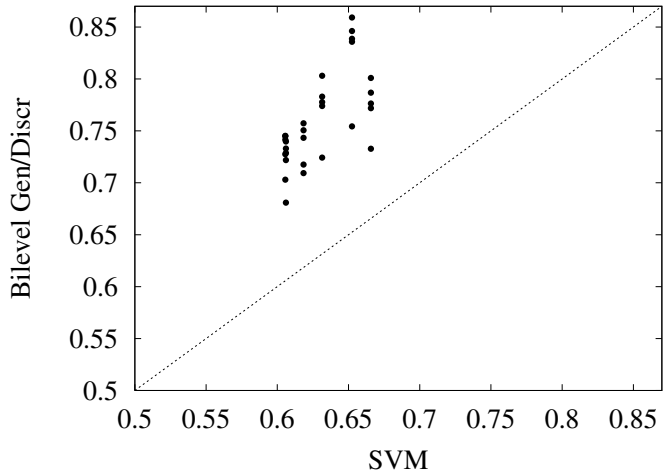}
\end{center}

\caption{ Performance of the bilevel discriminative classifier constrained by generative prior knowledge
versus performance of SVM. Each point represents a unique pair of training/test tasks, with 0.5\% of the test task data
used for training. The results are averaged over 100 experiments.}
\label{experiment_all}
\end{figure}

\begin{figure}
\begin{center}
\begin{minipage}[t]{0.33\textwidth}
\begin{flushleft}

{\footnotesize ~~~~~~1) Train:atheism vs. guns }

{\footnotesize ~~~~~~~~~~Test:guns vs. mideast}
\end{flushleft}

\includegraphics[%
  scale=0.35]{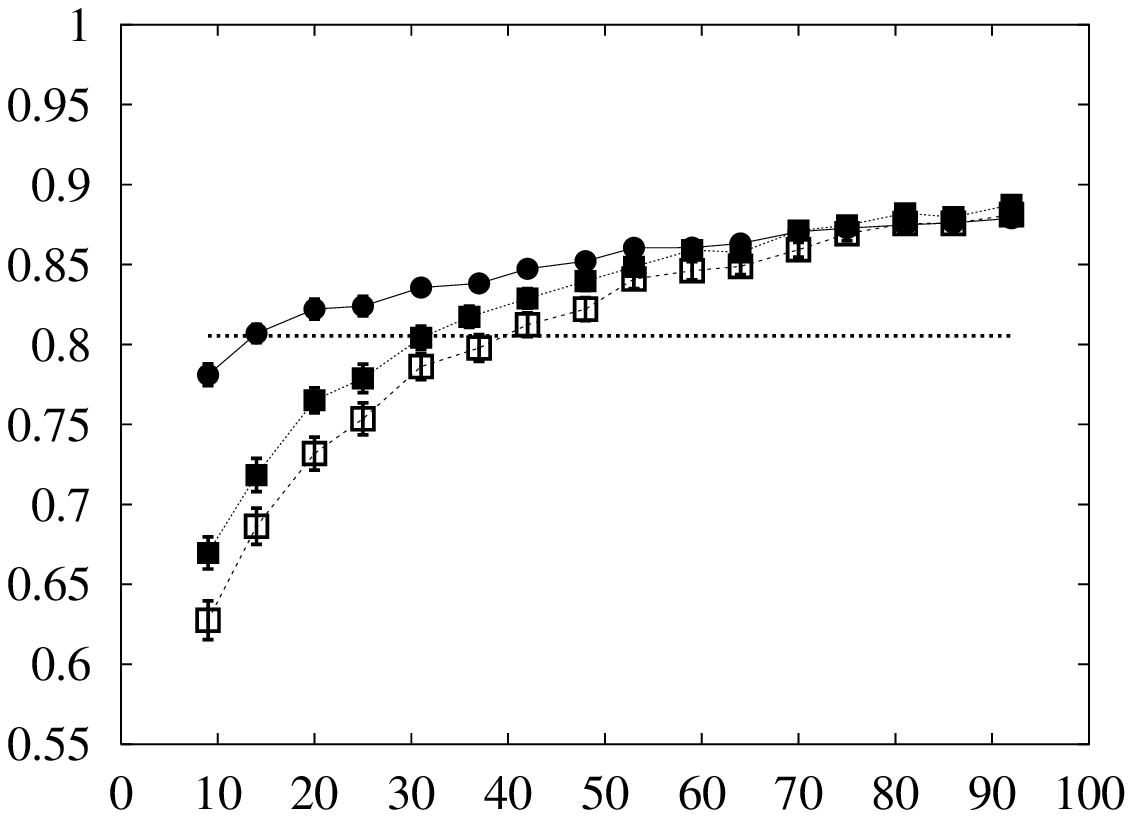}
\end{minipage}\begin{minipage}[t]{0.33\textwidth}
\begin{flushleft}
{\footnotesize ~~~~~~2) Train:atheism vs. guns} 

{\footnotesize ~~~~~~~~~~Test:atheism vs. mideast}
\end{flushleft}

\includegraphics[%
  scale=0.35]{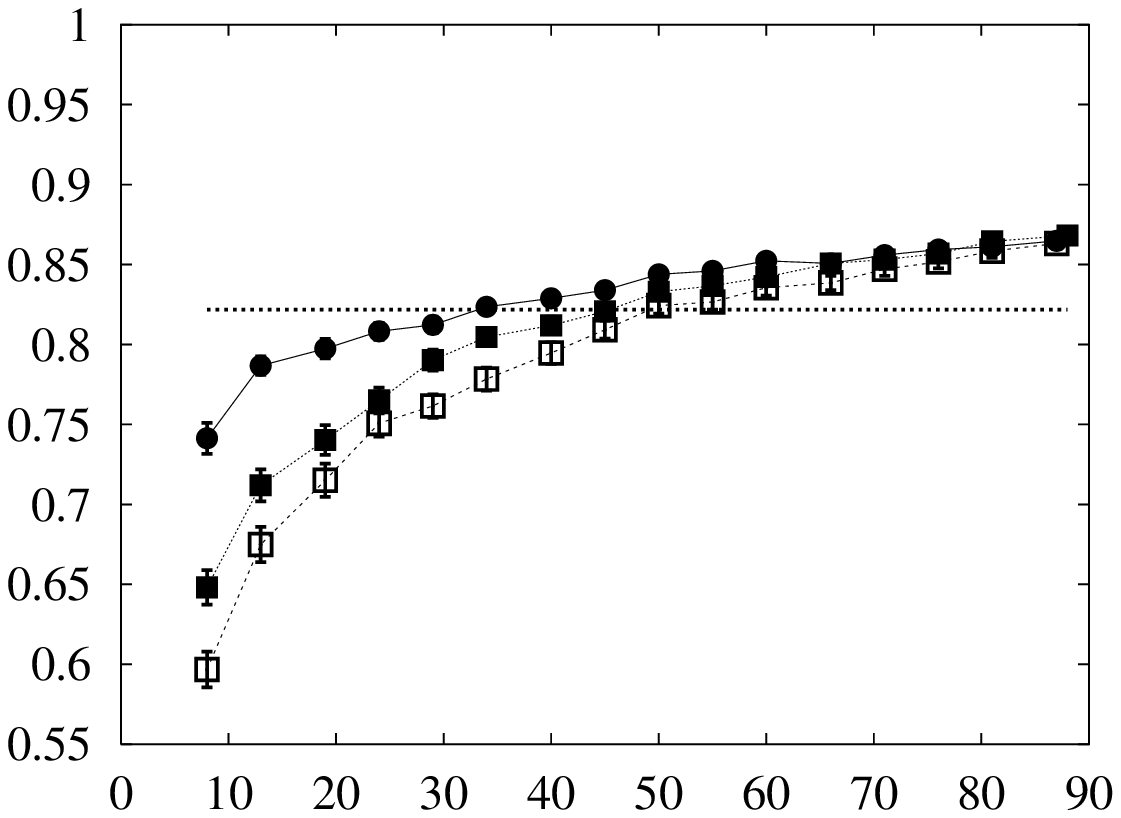}
\end{minipage}\begin{minipage}[t]{0.33\textwidth}
\begin{flushleft}
{\footnotesize ~~~~~~3) Train:guns vs. mideast}

{\footnotesize ~~~~~~~~~~Test:atheism vs. guns}
\end{flushleft}

\includegraphics[%
  scale=0.35]{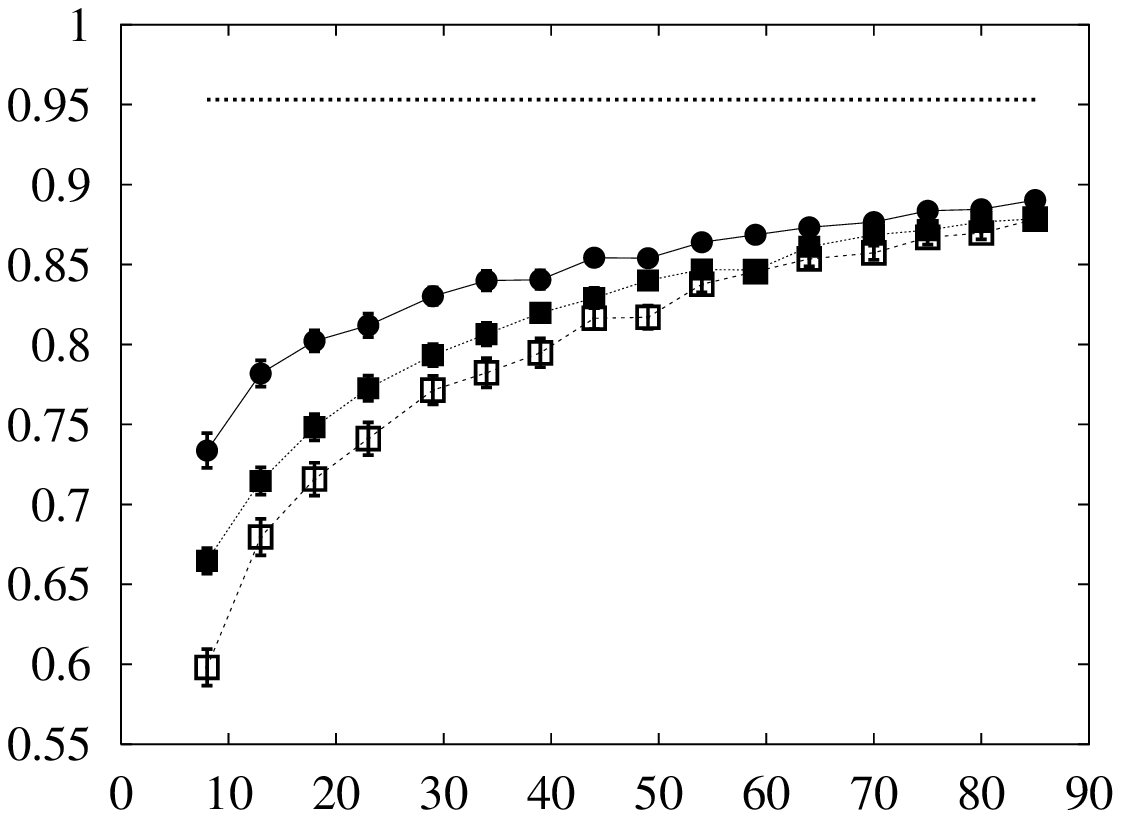}
\end{minipage}
\end{center}

\begin{center}
\begin{minipage}[c]{0.33\textwidth}
\begin{flushleft}
{\footnotesize ~~~~~~4) Train: guns vs. mideast} 

{\footnotesize ~~~~~~~~~~Test:atheism vs. mideast}
\end{flushleft}

\includegraphics[%
  scale=0.35]{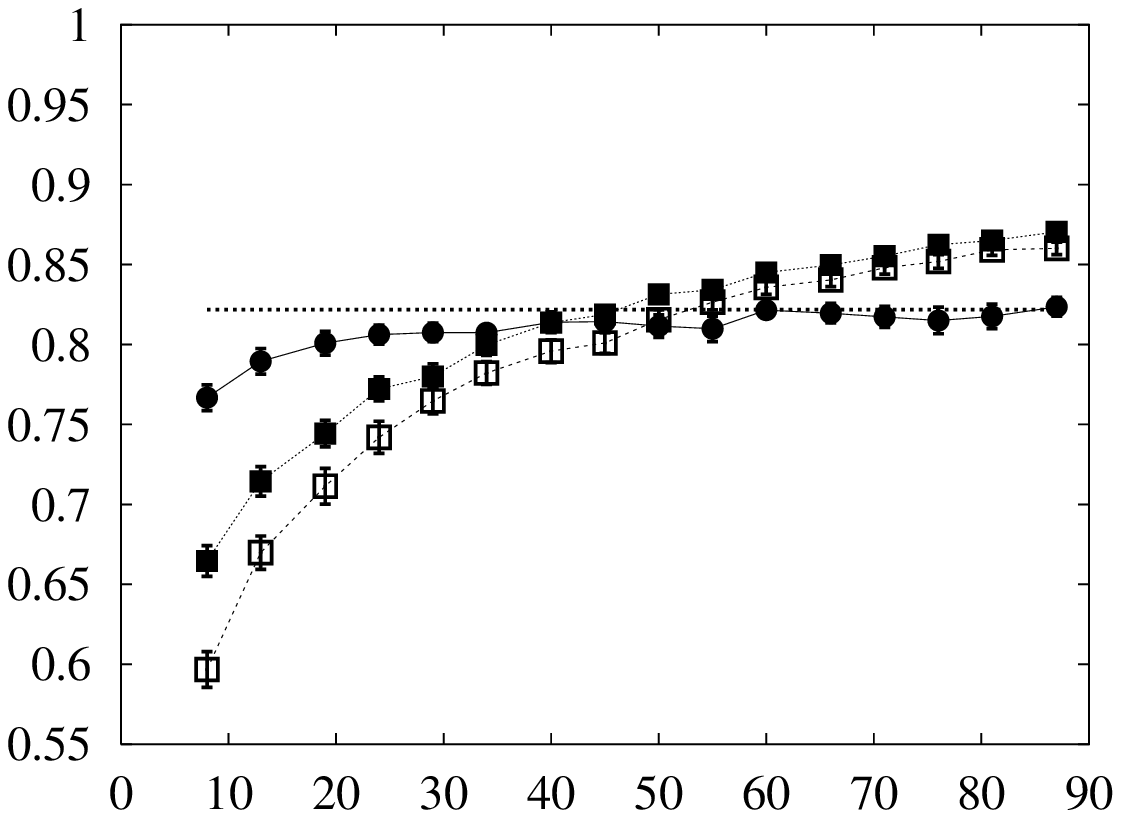}
\end{minipage}\begin{minipage}[c]{0.33\textwidth}
\begin{flushleft}
{\footnotesize ~~~5) Train: atheism vs. mideast} 

{\footnotesize ~~~~~~Test:guns vs. mideast}
\end{flushleft}

\includegraphics[%
  scale=0.35]{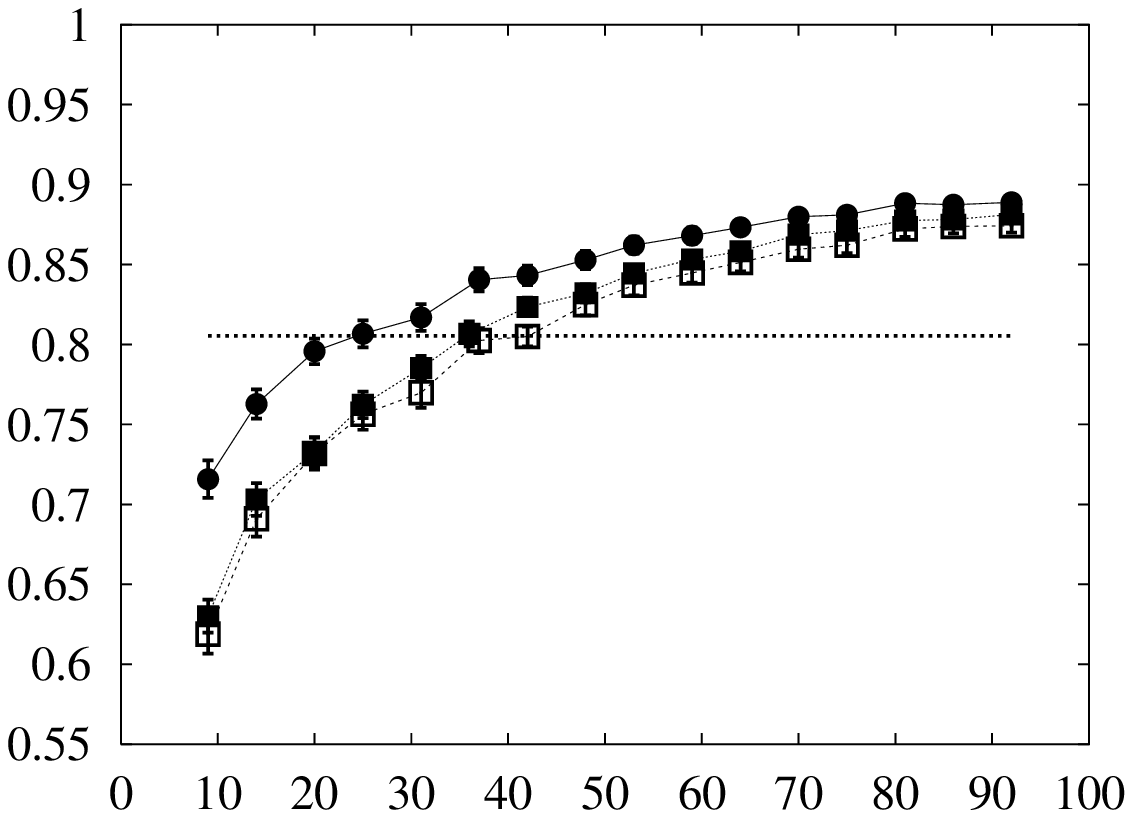}
\end{minipage}\begin{minipage}[c]{0.33\textwidth}
\begin{flushleft}
{\footnotesize ~~6) Train: atheism vs. mideast} 

{\footnotesize ~~~~~~Test:atheism vs. guns}
\end{flushleft}

\includegraphics[%
  scale=0.35]{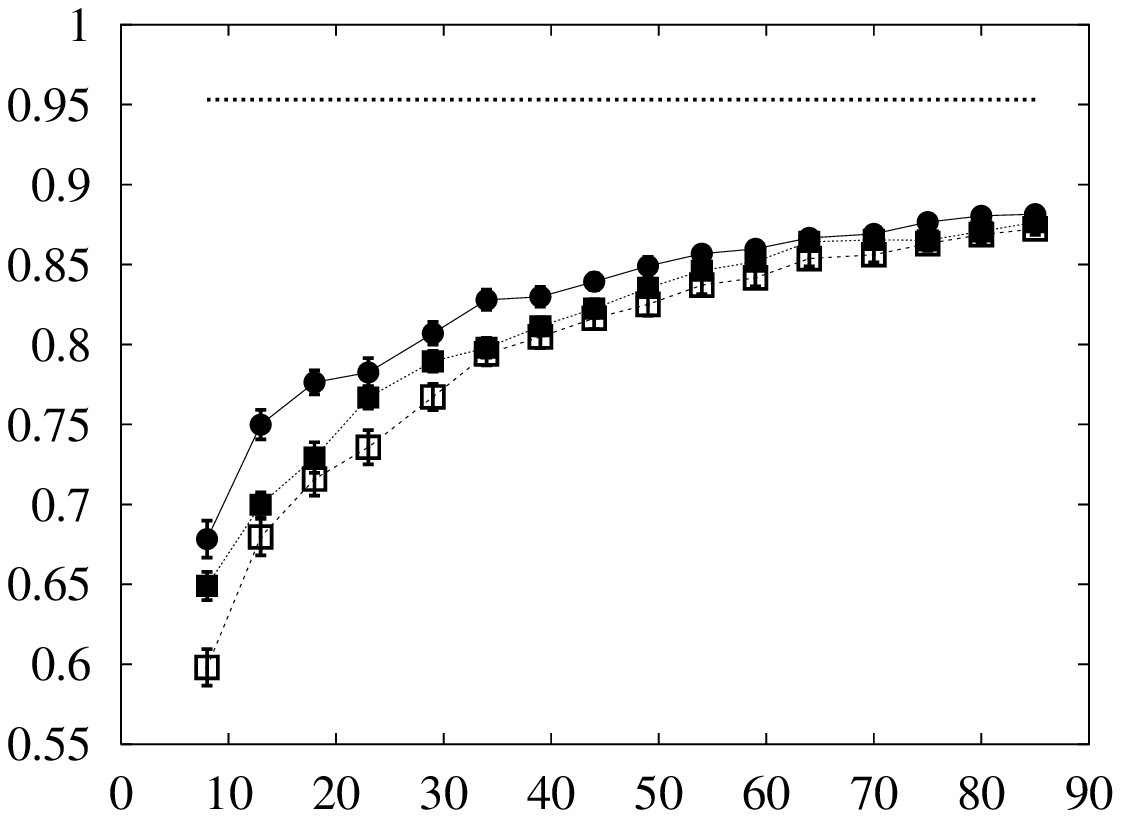}
\end{minipage}

\begin{minipage}[c]{.33\textwidth}
\includegraphics[%
  scale=0.35]{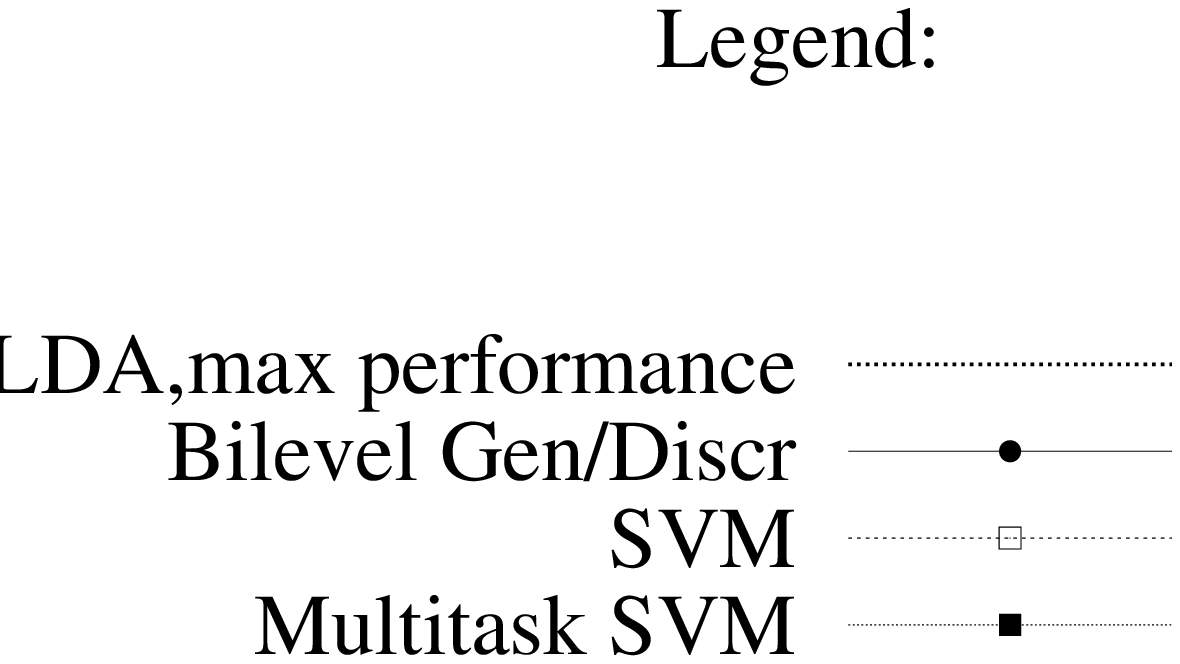}
\end{minipage}
\end{center}

\caption{ Test set accuracy as a percentage versus number of test task
training points for two classifiers (SVM and Bilevel Gen/Discr)
tested on six different classification tasks. For
each classification experiment, the data set was split randomly into
training and test sets in 100 different ways. 
The error bars based on 95\% confidence intervals.}
\label{experiment1}
\end{figure}

The experiments were designed both to demonstrate the usefulness of
the proposed approach for incorporation of generative prior into discriminative
classification, and to address a broader question by showing that
it is possible to use an existing domain theory to aid in a classification
task for which it was not specifically designed. In order to construct
the generative prior, the generative LDA classifier was trained on
the data from the training classification task to estimate the Gaussian
location parameters $\widehat{\mu_{i}}, i=1,2$, as described in Section 2.  The compression function $\chi_{i}(v)$ is subsequently computed (also as described in Section 2), and is used to set the hyperprior parameters via $\mu_{i}^{j}:=\chi_{i}(l^{j}_{i,test}),i=1,2$.  In order to apply a domain theory
effectively to the task for which it was not specifically designed, the algorithm must be able to estimate its confidence in the decomposition of the domain theory with respect to this new learning task. In order to model the uncertainty in applicability of WordNet to newsgroup categorization,
our system estimated its confidence in homogeneity of equivalence classes of semantic distances by computing the variance of each random variable $\chi_{i}(v)$ as follows: $\sigma_{i}(v)\triangleq \frac{\sum_{j:l^{j}_{i,train=v}}(\widehat{\mu_{i}^{j}}-\chi_{i}(v))^{2}}{|j:l^{j}_{i,tran}=v|}$.  The hyperprior confidence matrices $\Omega_{i},i=1,2$ were then reconstructed with respect to the test task semantic distances $l_{i,test},i =1,2$ as follows: 
$[\Omega_{i}]_{j,k}:=\left\{
\begin{array}{c}
\sigma_{i}(l^{j}_{i,test}),k=j\\
0,k\neq j
\end{array}
\right.
$.
Identity matrices were used
as covariance matrices of the lower-level prior: $\Sigma_{1}=\Sigma_{2}:=I$.
The rest of the parameters were set as follows: $\beta:=0.2,\varphi:=0.01$,
$C_{1}=C_{2}:=1$, $C_{3}:=\infty$. These constants were chosen manually
to optimize performance on Experiment 1 (for the training task:
atheism vs. guns, test task: guns vs. mideast, see Figure \ref{experiment1}) without observing any
data from any other classification tasks.

The resulting classifier was evaluated in different experimental setups (with different pairs of newsgroups chosen for the training and the test tasks) to justify the following claims:
\begin{enumerate}
\item The bilevel generative/discriminative classifier with WordNet-derived prior knowledge has good low-sample performance,
showing both the feasibility of automatically interpreting the knowledge embedded in WordNet and the efficacy of the proposed algorithm.
\item The bilevel classifier's performance improves with increasing training sample size.
\item Integrating generative prior into the discriminative classification framework results in better performance than 
integrating the same prior directly into the generative framework via Bayes' rule.
\item The bilevel classifier outperforms a state-of-the-art discriminative multitask classifier proposed by Evgeniou and Pontil \citeyear{multitask} by taking advantage of the WordNet domain theory.
\end{enumerate}
In order to evaluate the low-sample performance of the proposed classifier, four newsgroups from the 20-newsgroup dataset were selected for experiments: \emph{atheism, guns,
middle east}, and \emph{auto}.  Using these categories, thirty experimental setups were created 
for all the possible ways of assigning newsgroups to training and test tasks (with a 
pair of newsgroups assigned to each task, under the constraint that the training and test
pairs cannot be identical)\footnote{Newsgroup articles were preprocessed by removing words which could not be interpreted as nouns by WordNet.  This preprocessing ensured that only one part of WordNet domain theory was exercised and resulted in virtually no reduction in classification accuracy.}.
In each experiment, we compared the following two classifiers:
\begin{enumerate}
\item Our bilevel generative-discriminative classifier with the knowledge transfer functions $\chi_{i}(v), \sigma_{i}(v),i=1,2$
learned from the labeled training data provided for the training task (using 90\% of all the available data for that task). The resulting prior was
subsequently introduced into the discriminative classification framework
via our approximate bilevel programming approach
\item A vanilla SVM classifier which minimizes the regularized empirical risk:
\begin{align}
\min_{w,b,\xi_{i}} & \sum_{i=1}^{m}\xi_{i}+C_{1}\left \Vert w \right \Vert^{2}\\
\such & y_{i}(w^{T}x_{i}+b) \geq 1-\xi_{i},i=1,..,m
\end{align}
\end{enumerate}

Both classifiers were trained on 0.5\% of all the available data from the \emph{test} classification task\footnote{SeDuMi software 
\cite{sturm-sedumi} 
was used to solve the iterative
SOCP programs.%
}, and evaluated on the remaining 99.5\% of the test task data.  The results, averaged over one hundred randomly selected datasets, are presented in Figure \ref{experiment_all},
which shows the plot of the accuracy of the bilevel generative/discriminative classifier versus the accuracy of the SVM classifier, 
evaluated in each of the thirty experimental setups.  All the points lie above the 45$^o$ line, indicating 
improvement in performance due to incorporation of prior knowledge via the bilevel programming framework.  
The amount of improvement ranges from 10\% to 30\%, with all of the improvements being statistically significant at the 5\% level.

The next experiment was conducted to evaluate the effect of increasing training data (from the test task) on the performance of the system.  For this experiment, we selected three newsgroups (\emph{atheism, guns,} and \emph{middle east}) and generated six experimental setups based
on all the possible ways of splitting these newsgroups into unique training/test pairs.   In addition to the classifiers 1 and 2 above, the following classifiers were evaluated:

\begin{enumerate}
\item[3.] A state-of-the art multi-task classifier designed by Evgeniou and Pontil \citeyear{multitask}.  The classifier learns a set of related classification functions $f_{t}(x)=w_{t}^{T}x+b_{t}$ for classification tasks $t \in \{$training task, test task$\}$ given $m(t)$ data points $[x_{1t},y_{1t}], .., [x_{m(t)t},y_{m(t)t}]$ for each task $t$ by minimizing the regularized empirical risk:
\begin{align}
\min_{w_{0},w_{t},b_{t},\xi_{it}}  \sum_{t}\sum_{i=1}^{m(t)}\xi_{it}+\frac{C_{1}}{C_{2}}\sum_{t}\left \Vert w_{t}-w_{0} \right \Vert^{2}+C_{1}\left \Vert w_{0} \right \Vert^{2} \\
\such  y_{it}(w_{t}^{T}x_{it}+b_{t}) \geq 1-\xi_{it} ,i=1,..,m(t),\forall t \\  \xi_{it} \geq 0,i=1,..,m(t),\forall t
\end{align}
The regularization constraint captures a tradeoff between final models $w_{t}$ being close to the average model $w_{0}$ and having a large margin on the training data.  90\% of the training task data was made available to the classifier.  Constant $C_{1}:=1$ was chosen, and $C_{2}:=1000$ was selected from the set $\{.1,.5,1,2,10,1000, 10^{5},10^{10}\}$ to optimize the classifier's performance on Experiment 1 (for the training task: atheism vs. guns, test task: guns vs. mideast, see Figure \ref{experiment1}) after observing .05\% of the test task data (in addition to the training task data).

\item[4.] The LDA classifier described in Section 2 trained on 90\% of the \emph{test} task data.  Since this classifier is the same as the bottom-level generative classifier used in the bilevel algorithm, its performance gives an upper bound on the performance of the bottom-level classifier trained in a generative fashion. 
\end{enumerate}

Figure \ref{experiment1} shows performance of classifiers 1-3 as a function of the size of the training data from the test task (evaluation was done on the remaining test-task data).  The results are averaged over one hundred randomly selected datasets.   The performance of the bilevel
classifier improves with increasing training data both because the discriminative portion of the classifier aims to minimize the training error and because
the generative prior is imposed with soft constraints.  As expected, the performance curves of the classifiers converge 
as the amount of available training data increases.  Even though the constants used in the mathematical program were selected in a single experimental setup,
the classifier's performance is reasonable for a wide range of data sets across different experimental setups, with the possible exception of 
Experiment 4 (training task: guns vs. mideast, testing task: atheism vs. mideast), where 
the means of the constructed elliptical priors are much closer to each other than in the other experiments. Thus, the prior is imposed with greater confidence than 
is warranted, adversely affecting the classifier's performance.

The multi-task classifier 3 outperforms the vanilla SVM by generalizing from data points across classification tasks.  However, it does not take advantage of 
prior knowledge, while our classifier does. The gain in performance of the bilevel generative/discriminative classifier is due to the fact that the relationship between the classification tasks is captured much better by WordNet than by simple linear averaging of weight vectors.

Because of the constants involved in both the bilevel classifier and the generative classifiers with Bayesian priors,
it is hard to do a fair comparison between classifiers constrained by generative priors in these two frameworks. Instead, the generatively trained classifier 4 gives an empirical upper bound on the performance achievable by the 
bottom-level classifier trained generatively on the test task data.
The accuracy of this classifier is shown as 
as a horizontal in the plots in Figure \ref{experiment1}.  Since discriminative classification is known to be superior to generative classification for this problem,
the SVM classifier outperforms the generative classifier given enough data in four out of six experimental setups.  What is more interesting, is that, for
a range of training sample sizes, the bilevel classifier constrained by the generative prior outperforms both the SVM trained on the same 
sample and the generative classifier trained on a much larger sample in these four setups. This means that, unless prior knowledge outweighs 
the effect of learning, it cannot enable the LDA classifier to compete with our bilevel classifier on those problems.

\begin{figure}
\begin{center}
\begin{minipage}[t]{0.5\textwidth}
\begin{center}

1) Accuracy as a function of $\beta$

\includegraphics[%
  scale=0.35]{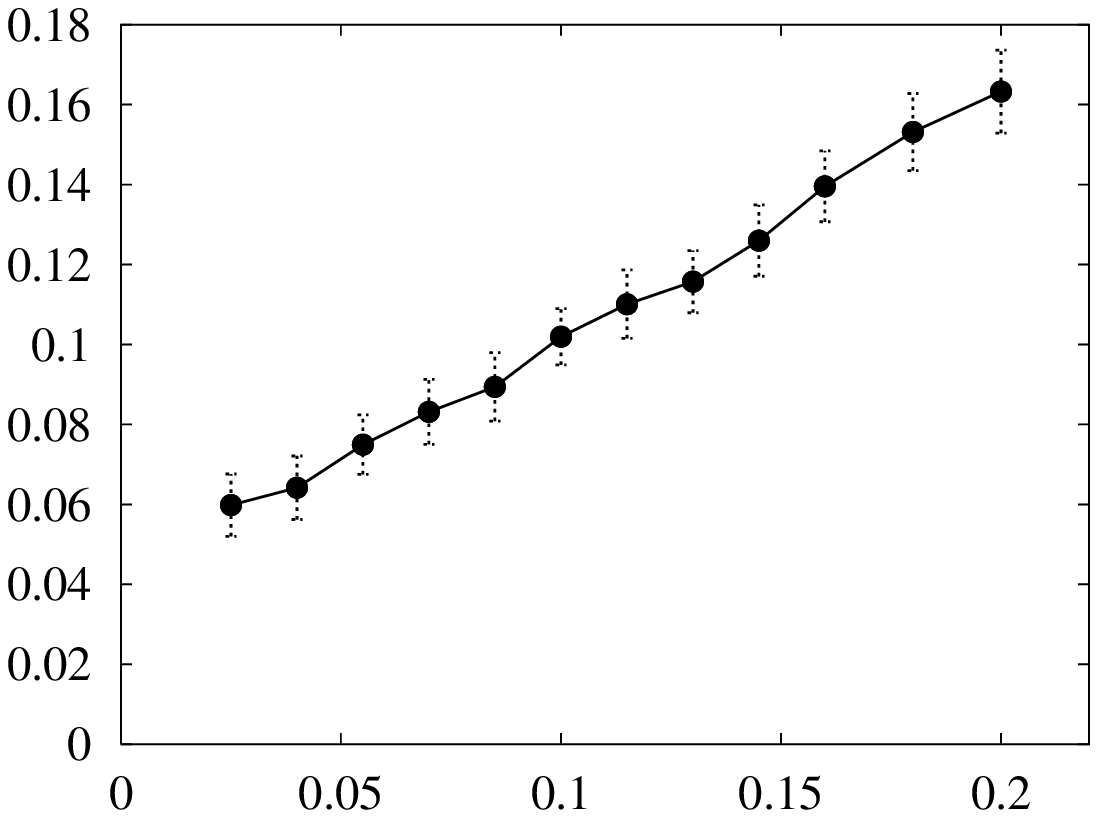}
\end{center}
\end{minipage}\begin{minipage}[t]{0.5\textwidth}
\begin{center}

2) Accuracy as a function of $\varphi$

\includegraphics[%
  scale=0.35]{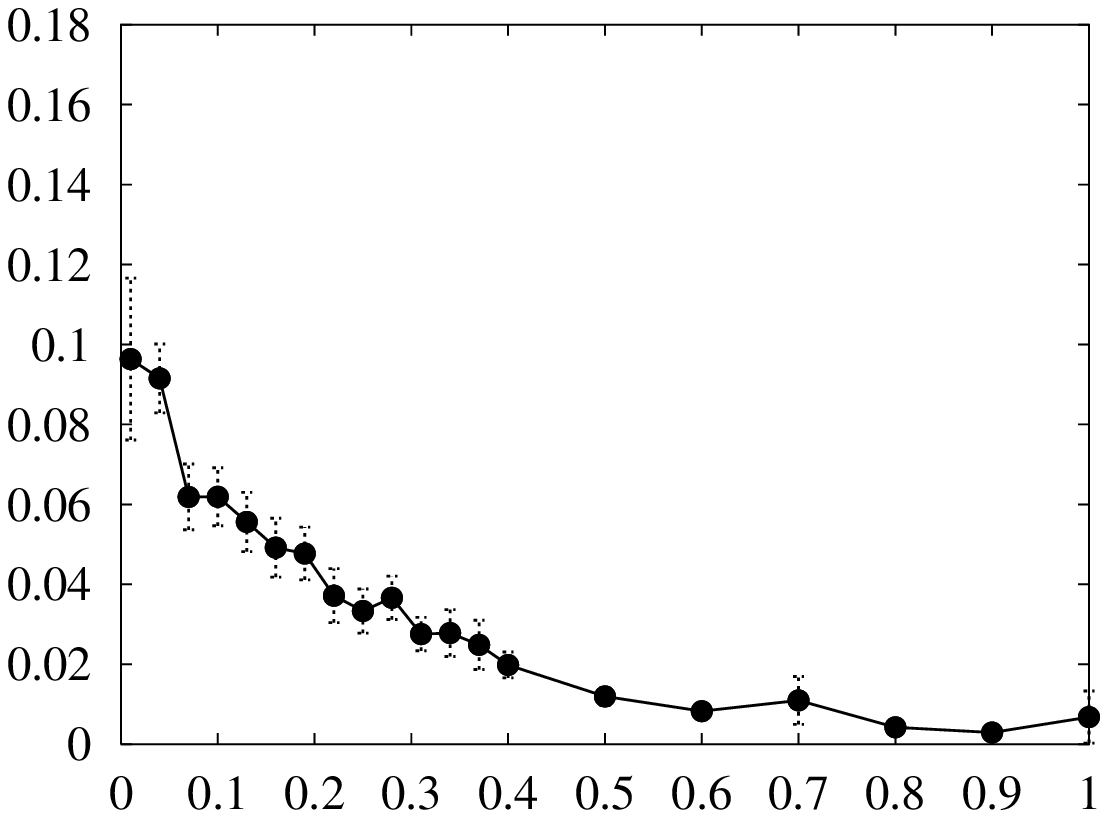}
\end{center}

\end{minipage}
\end{center}

\caption{Plots of test set accuracy as percentage versus mathematical program parameter values. 
For each classification task, a random training set of size 9 was chosen from the
full set of test task articles in 100 different ways. Error bars are based on 95\% confidence intervals.
All the experiments were performed on the training task: atheism vs. guns, test task: guns vs. mideast.
}
\label{sensitivity}
\end{figure}

Finally, a set of experiments was performed to determine the effect of varying mathematical program
parameters $\beta$ and $\varphi$ on the generalization error.  Each parameter was varied over a set of values,
with the rest of the parameters held fixed ($\beta$ was increased up to its maximum feasible value).  
The evaluation was done in the setup of Experiment 1 (for the training task:atheism vs. guns, test task: guns vs. mideast),
with the training set size of 9 points.  The results are presented in Figure \ref{sensitivity}. 
Increasing the value of $\beta$ is equivalent to requiring a hyperplane separator to have smaller error
given the prior. Decreasing the value of $\varphi$ is equivalent to increasing the confidence in the hyperprior.
Both of these actions tighten the constraints (i.e., decrease the feasible region).  With good prior knowledge, 
this should have the effect of improving generalization performance for small training samples since 
the prior is imposed with higher confidence.  This is precisely what we observe in the plots of Figure \ref{sensitivity}.

\section{Generalization Performance}

Why does the algorithm generalize well for low sample sizes? In this
section, we derive a theorem which demonstrates that the convergence
rate of the generalization error of the constrained generative-discriminative
classifier depends on the parameters of the mathematical program and
not just the margin, as would be expected in the case of large-margin
classification without the prior. In particular, we show that as the certainty of 
the generative prior knowledge increases, the upper bound on the 
generalization error of the classifier constrained by the prior decreases. 
By increasing certainty of the prior, we mean that either the hyper-prior
becomes more peaked (i.e., the confidence in the locations of the prior means
increases) or the desired upper bounds on the Type I and Type II probabilities of error of 
the classifier decrease (i.e., the requirement that the lower-level discriminative player 
choose the restricted Bayes-optimal hyperplane is more strictly enforced).

The argument proceeds by bounding 
the fat-shattering dimension of the classifier constrained by prior
knowledge. The fat-shattering dimension of a large margin classifier
is given by the following definition \cite{bartlett:generalize}: 

\begin{defn}
\emph{A set of points $S=\{ x^{1}...x^{m}\}$ is $\gamma$-shattered
by a set of functions $F$ mapping from a domain $X$ to $\mathbb{{R}}$
if there are real numbers $r^{1},...,r^{m}$ such that, for each $b\in\{-1,1\}^{m}$,
there is a function $f_{b}$ in $F$ with $b(f_{b}(x^{i})-r^{i})\geq\gamma,\, i=1..m$.
We say that $r^{1},...,r^{m}$ witness the shattering. Then the fat-shattering
dimension of $F$ is a function fat$_{F}$($\gamma$) that maps $\gamma$
to the cardinality of the largest $\gamma$-shattered set $S$.} 
\end{defn}
Specifically, we consider the class of functions 
\begin{equation}
\label{eqnf}
F=\{ x\rightarrow w^{T}x:\left\Vert x\right\Vert \leq R,\left\Vert w\right\Vert =1,
\end{equation}
\begin{equation*}
%\notag
%\\
\frac{w^{T}(-\mu_{1})}{\left\Vert \Sigma_{1}^{1/2}w\right\Vert }\geq\beta,
\left\Vert \Omega_{1}^{-1/2}(\mu_{1}-t_{1})\right\Vert \leq\varphi,
\frac{w^{T}\mu_{2}}{\left\Vert \Sigma_{2}^{1/2}w\right\Vert }\geq\beta,
\left\Vert \Omega_{2}^{-1/2}(\mu_{2}-t_{2})\right\Vert \leq\varphi\}. 
\end{equation*}
%\end{align}

The following theorem bounds the fat-shattering dimension of our classifier:
\begin{thm}
\label{thm-gen-bound}
Let $F$ be the class of a-priori constrained functions defined by (\ref{eqnf}),
and let $\lambda_{min}(P)$ and $\lambda_{max}(P)$ denote the minimum and maximum eigenvalues of matrix $P$, respectively.
If a set of points $S$ is $\gamma$-shattered by $F$, then
$\left|S\right|\leq\frac{4R^{2}(\alpha^{2}(1-\alpha^{2}))}{\gamma^{2}}$,
where $\alpha=\max(\alpha_{1},\alpha_{2})$ with $\alpha_{1}=\min(\frac{\lambda_{min}(\Sigma_{1})\beta}{\left\Vert \mu_{2}\right\Vert },\frac{\left\Vert t_{2}\right\Vert ^{2}-(\lambda_{max}(\Omega_{2})\varphi)^{2}}{\left\Vert t_{2}\right\Vert (\lambda_{max}(\Omega_{2})\varphi)^{2}+\left\Vert t_{2}\right\Vert )})$
and $\alpha_{2}=\min(\frac{\lambda_{min}(\Sigma_{1})\beta}{\left\Vert \mu_{1}\right\Vert },$
$\frac{\left\Vert t_{1}\right\Vert ^{2}-(\lambda_{max}(\Omega_{1})\varphi)^{2}}{\left\Vert t_{1}\right\Vert ((\lambda_{max}(\Omega_{1})\varphi)^{2}+\left\Vert t_{1}\right\Vert )})$,
assuming that $\beta\geq0$, $\left\Vert t_{i}\right\Vert \geq\left\Vert t_{i}-\mu_{i}\right\Vert $,
and $\alpha_{i}\geq\frac{1}{\sqrt{2}},i=1,2$. 
\begin{proof}
See Appendix \ref{appendix-gen-bound}.
\end{proof}
\end{thm}

We have the following corollary which follows directly from Taylor and Bartlett's \citeyear{bartlett:generalize} Theorem
1.5 and bounds the classifier's generalization
error based on its fat-shattering dimension:

\begin{cor}
Let $G$ be a class of real-valued functions. Then, with probability
at least $1-\delta$ over $m$ independently generated examples $z$,
if a classifier $h=sgn(g)\in sgn(G)$ has margin at least $\gamma$
on all the examples in $z$, then the error of $h$ is no more than
$\frac{2}{m}(d*log(\frac{8em}{d})log(32m)+log(\frac{8m}{\delta}))$
where $d_{G}=fat_{G}(\frac{\gamma}{16})$. If $G=F$ is the class
of functions defined by (\ref{eqnf}), then $d_{F}\leq\frac{265R^{2}(4(\alpha^{2}(1-\alpha^{2})))}{\gamma^{2}}$.
If $G=F'$ is the usual class of large margin classifiers (without
the prior), then the result in \cite{bartlett:generalize} shows that
$d_{F'}\leq\frac{265R^{2}}{\gamma^{2}}$.
\end{cor}
Notice that both bounds depend on $\frac{R^{2}}{\gamma^{2}}$. However,
the bound of the classifier constrained by the generative prior also
depends on $\beta$ and $\varphi$ through the term $4(\alpha^{2}(1-\alpha^{2}))$.
In particular, as $\beta$ increases, tightening the constraints,
the bound decreases, ensuring, as expected, quicker convergence of
the generalization error. Similarly, decreasing $\varphi$ also tightens
the constraints and decreases the upper bound on the generalization
error. For $\alpha > \frac{1}{\sqrt{2}}$, the factor $4(\alpha^{2}(1-\alpha^{2}))$
is less than $1$ and the upper bound on the fat-shattering dimension
$d_{F}$ is tighter than the usual bound in the no-prior case on $d_{F'}$.

Since $\beta$ controls the amount of deviation of the decision boundary from the 
Bayes-optimal hyperplane and $\varphi$ depends on the variance of the hyper-prior distribution, tightening of
these constraints corresponds to increasing our confidence in the prior. Note that a high value $\beta$
represents high level of user confidence in the generative elliptical model.
Also note that there are two ways of 
increasing the tightness of the hyperprior constraint (\ref{eqn_means}) - one is through the user-defined parameter  
$\varphi$, the other is through the automatically estimated covariance matrices $\Omega_{i},i=1,2$.
These matrices estimate the extent to which the equivalence classes defined by WordNet create
an appropriate decomposition of the domain theory for the newsgroup categorization task.  Thus,
tight constraint (\ref{eqn_means}) represents both high level of user confidence in the means of the generative classification model
 (estimated from WordNet) and a good correspondence between the partition of the words imposed by the semantic distance of WordNet
and the elliptical generative model of the data. As $\varphi$ approaches zero and $\beta$ approaches its highest feasible value,
the solution of the bilevel mathematical program reduces to the restricted Bayes optimal decision boundary computed solely
from the generative prior distributions, without using the data. 

Hence, we have shown that, as the prior is imposed with increasing level of confidence (which
means that the elliptical generative model is deemed good, or the estimates of its means are good, which in turn implies that
the domain theory is well-suited for the classification task at hand), the convergence rate of the generalization error 
of the classifier increases.  Intuitively, this is precisely the desired effect of increased confidence in the prior since 
the benefit derived from the training data is outweighed by the benefit derived from prior knowledge. 
For low data samples, this should result in improved accuracy assuming the domain theory is good, which is what the 
 plots in Figure \ref{sensitivity} show.

\section{Related Work}
There are a number of approaches to combining generative and discriminative models.
%\cite{Tong:restrictedbayes,t-rvm,bn-discr,bn-logr,jebara-book,jaakkolla:maxent,ng:hybrid,ng:logregr,scholkopf,fung,wu,rotate-svm,qiang-ebl,mangasarian,hmm-perceptron}.
Several of these focus on deriving discriminative classifiers
from generative distributions 
\cite{Tong:restrictedbayes,t-rvm} 
or
on learning the parameters of generative classifiers via discriminative
training methods 
\cite{bn-discr,bn-logr}. 
The closest in spirit to our approach is the Maximum Entropy Discrimination
framework \cite{jebara-book,jaakkolla:maxent}, which performs
discriminative estimation of parameters of a generative model, taking into
account the constraints of fitting the data and respecting the prior. 
One important difference with our framework is that, in estimating these parameters, maximum entropy discrimination minimizes the distance between the generative model and the prior, subject to satisfying the discriminative constraint that the training data be classified correctly with a given margin.  Our framework, on the other hand, maximizes the margin on the training data subject to the constraint that the generative model is not too far from the prior.  This emphasis on maximizing the margin allows us to derive a-priori bounds on the generalization error of our classifier based on the confidence in the prior which are not (yet) available for the maximum entropy framework.  Another difference is that our approach performs classification via a single generative model, while maximum entropy discrimination averages over a set of generative models weighted by their probabilities.  This is similar to the distinction between maximum-a-posteriori and Bayesian estimation and has repercussions for tractability.  Maximum entropy discrimination, however, is more general than our framework in a sense of allowing a richer set of behaviors based on different priors.

Ng et al. \citeyear{ng:hybrid,ng:logregr} explore the relative advantages of discriminative and generative classification
and propose a hybrid approach which improves classification accuracy
for both low-sample and high-sample scenarios. 
Collins \citeyear{hmm-perceptron} proposes to use
the Viterbi algorithm for HMMs for inferencing (which is based on generative assumptions),
combined with a discriminative learning algorithm for HMM parameter estimation.
These research directions are orthogonal to our work since they do not explicitly consider the question of integration of prior knowledge into the learning problem.

In the context of support vector classification, various forms of
prior knowledge have been explored. Scholkopf et al. \citeyear{scholkopf}
demonstrate how to integrate prior knowledge about invariance under
transformations and importance of local structure into the kernel
function. Fung et al. \citeyear{fung} use domain knowledge in form of
labeled polyhedral sets to augment the training data. Wu and Srihari
\citeyear{wu} allow domain experts to specify their confidence in the example's
label, varying the effect of each example on the separating hyperplane
proportionately to its confidence. Epshteyn and DeJong \citeyear{rotate-svm}
explore the effects of rotational constraints on the normal of the
separating hyperplane. Sun and DeJong \citeyear{qiang-ebl} propose an algorithm which uses domain knowledge (such as WordNet) to identify relevant features
of examples and incorporate resulting information in form of soft constraints
on the hypothesis space of SVM classifier.  Mangasarian et al. \citeyear{mangasarian} suggest the
use of prior knowledge for support vector regression. In all of these
approaches, prior knowledge takes the form of explicit constraints
on the hypothesis space of the large-margin classifier. In this work,
the emphasis is on generating such constraints automatically from
domain knowledge interpreted in the generative setting. As we demonstrate
with our WordNet application, generative interpretation of background
knowledge is very intuitive for natural language processing problems.

Second-order cone constraints have been applied extensively to model
probability constraints in robust convex optimization \cite{lobo1,nips:missingdata}
and constraints on the distribution of the data in minimax machines
 \cite{minimax1,minimax2004}. Our work, as far as we know, is the
first one which models prior knowledge with such constraints. The
resulting optimization problem and its connection with Bayes optimal
classification is very different from the approaches mentioned above. 

Our work is also related to empirical Bayes estimation \cite{empirical-bayes}.
In empirical Bayes estimation, the hyper-prior parameters of the generative
model are estimated using statistical estimation methods (usually
maximum likelihood or method of moments) through the marginal distribution
of the data, while our approach learns those parameters discriminatively
using the training data.

\section{Conclusions and Future Work.}

Since many sources of domain knowledge (such as WordNet) are readily available, we believe that significant benefit can be achieved by developing algorithms for automatically applying their information to new 
classification problems.  In this paper, we argued that the generative paradigm for interpreting
background knowledge is preferable to the discriminative interpretation, and presented a novel algorithm
which enables discriminative classifiers to utilize generative prior knowledge.  Our algorithm was 
evaluated in the context of a complete system which, faced with the newsgroup classification task,
was able to estimate the parameters needed to construct the generative prior from the domain theory, and use
this construction to achieve improved performance on new newsgroup classification tasks.  

In this work, we restricted our hypothesis class to that of linear classifiers.
Extending the form of the prior distribution to distributions other than elliptical
and/or looking for Bayes-optimal classifiers restricted to a more expressive class
than that of linear separators may result in improvement in classification accuracy for 
non linearly-separable domains.  However, it is not obvious how to approximate this more expressive
form of prior knowledge with convex constraints.  The kernel trick may be helpful in 
handling nonlinear problems, assuming that it is possible to represent the
optimization problem exclusively in terms of dot products of the data points and constraints.
This is an important issue which requires further study.

We have demonstrated that interpreting domain theory in the generative setting 
is intuitive and produces good empirical results. However, there are usually multiple 
ways of interpreting a domain theory.  In WordNet, 
for instance, semantic distance between words is only one measure of information
contained in the domain theory.  Other, more complicated, interpretations might, for example, take into
account types of links on the path between the words (hypernyms, synonyms, meronyms,
etc.) and exploit common-sense observations about WordNet such
as words that are closer to the category label are more likely to
be informative than words farther away. Comparing multiple ways of constructing
the generative prior from the domain theory and, ultimately, 
selecting one of these interpretations automatically is a fruitful direction for further research.

\acks{The authors thank the anonymous reviewers for valuable suggestions on improving the paper.  This material is based upon work supported in part 
by the National 
Science Foundation under Award  NSF IIS 04-13161 and in part by the 
Information Processing Technology Office of the Defense Advanced 
Research Projects Agency under award HR0011-05-1-0040. Any opinions, 
findings, and conclusions or recommendations expressed in this 
publication are those of the authors and do not necessarily reflect the 
views of the National Science Foundation or the Defense Advanced 
Research Projects Agency.}

\appendix
\section{Convergence of the Generative/Discriminative Algorithm}
\label{appendix-dual}
Let the map $H:Z \rightarrow Z$ determine an algorithm that,
given a point $\mu^{(0)}$, generates a sequence $\left\{\mu^{(t)}\right\}_{t=0}^{\infty}$ of iterates through the iteration $\mu^{(t+1)} = H(\mu^{(t)})$.
The iterative algorithm in Section \ref{ch-gen-discr} generates a sequence of 
iterates $\mu^{(t)}=[\mu_{1}^{(t)},\mu_{2}^{(t)}]\in Z$ by applying the following map $H$: 
\begin{equation}
\label{eqn_alg_map}
H=H_{2}\circ H_{1}:
\end{equation}
\begin{align}
\text{In step 1, }H_{1}([\mu_{1},\mu_{2}])=\arg\min_{[w,b] \in U([\mu_{1},\mu_{2}])}\left\Vert w \right \Vert,\label{eqn_alg_map1}\\
\text{with the set }U([\mu_{1},\mu_{2}])\text{ defined by constraints: }\\
y_{i}(w^{T}x_{i}+b)-1\geq 0, i=1,..,m\label{eqn_constr_a1}\\
c_{-1}(w,b;\mu_{1},\Sigma_{1})-\beta\geq 0\label{eqn_constr_a2}\\
c_{1}(w,b;\mu_{2},\Sigma_{2})-\beta\geq 0\label{eqn_constr_a3}
\end{align}
%\text{ given by constraints (\ref{eqn_constr1})-(\ref{eqn_constr2})}\notag

with the conic constraints $c_{s}(w,b;\mu,\Sigma)\triangleq s\left(\frac{w^{T}\mu+b}{\left\Vert \Sigma^{1/2}w \right\Vert}\right)$.

\begin{align}
\text{In step 2, }
H_{2}(w,b)=\arg\min_{(\mu_{1},\mu_{2}) \in V}-(c_{-1}(w,b;\mu_{1},\Sigma_{1})+c_{1}(w,b;\mu_{2},\Sigma_{2}))
%\frac{w^{T}\mu_{2}+b}{\left\Vert \Sigma_{2}^{1/2}w\right\Vert }-\frac{w^{T}\mu_{1}+b}{\left\Vert \Sigma_{1}^{1/2}w\right\Vert },
\label{eqn_alg_map2}\\
\text{with the set }V\text{ given by the constraints }\notag\\
\varphi-o(\mu_{1};\Omega_{1},t_{1})\geq 0 \label{eqn_constr_b1}\\
\varphi-o(\mu_{2};\Omega_{2},t_{2})\geq 0 \label{eqn_constr_b2}
%%\left\Vert \Omega_{i}^{-1/2}(\mu_{i}-t_{i})\right\Vert\geq 0 \notag
%\text{ given by the constraint (\ref{eqn_constr4})}\notag
\end{align}
with $o(\mu;\Omega,t)\triangleq \left\Vert \Omega^{-1/2}(\mu-t)\right\Vert$.

Notice that $H_{1}$ and $H_{2}$ are functions because the minima for optimization problems (\ref{eqn_opt_step1})-(\ref{eqn_constr3}) and (\ref{eqn_opt_step2})-(\ref{eqn_constr4}) are unique.  
%This is the case because the objectives and the constraints are convex, and the Lagrangians are strictly convex in the optimization parameters
This is the case because Step 1 optimizes a strictly convex function on a convex set, and Step 2 optimizes a linear non-constant function on a strictly convex set.
% (i.e., a set such that for any two points $z_{1},z_{2}$, points $z=\lambda z_{1}+(1-\lambda)z_{2}, \lambda \in (0,1)$ on the line connecting $z_{1}$ and $z_{2}$ lie in the interior of the set).

Convergence of the objective function $\psi(\mu^{(t)})\triangleq \min_{[w,b]\in U([\mu_{1}^{(t)},\mu_{2}^{(t)}])} \left \Vert w \right \Vert$ of the algorithm was shown in Theorem
\ref{thm-obj-conv}.  Let $\Gamma$ denote the set of points on which the map $H$ does not change the value of the objective function, i.e. $\mu^{*} \in \Gamma \Leftrightarrow \psi(H(\mu^{*}))=\psi(\mu^{*})$.
We will show that every accumulation point of $\{\mu^{(t)}\}$ lies in $\Gamma$.  
We will also show that every point $[\mu_{1}^{*},\mu_{2}^{*}]\in \Gamma$ augmented with $[w^{*},b^{*}]=H_{1}([\mu_{1}^{*},\mu_{2}^{*}])$ is a point with no feasible descent directions for the optimization problem (\ref{eqn_opt})-(\ref{eqn1_constr4}), which can be equivalently expressed as:
\begin{equation}
\min_{\mu_{1},\mu_{2},w,b} \left\Vert w \right\Vert 
\such [\mu_{1},\mu_{2}]\in V;[w,b]\in U([\mu_{1},\mu_{2}])
\label{eqn_orig_opt}
\end{equation}

In order to formally state our result, we need a few concepts from the duality theory.
Let a constrained optimization problem be given by 
\begin{equation}\min_{x}f(x) \such c_{i}(x)\geq 0, i=1,..,k
\label{eqn_constr_opt}
\end{equation}

%The Lagrangian for (\ref{eqn_constr_opt}) is defined as $L(x,\boldsymbol{\lambda}:=(\lambda_{1},..,\lambda_{k})):=f(x)-\sum_{i=1}^{k}\lambda_{i}c_{i}(x)$.
The following conditions, known as Karush-Kuhn-Tucker(KKT) conditions are necessary for $x^{*}$ to be a local minimum:
\begin{prop}
If $x^{*}$ is a local minimum of (\ref{eqn_constr_opt}), then $\exists \lambda_{1},..,\lambda_{k}$ such that
\begin{enumerate}
\item $\nabla f(x^{*})=\sum_{i=1}^{k}\lambda_{i}\nabla c_{i}(x^{*})$
\item $\lambda_{i}\geq 0 \text{ for } \forall i \in \{1,..,k\}$
\item $c_{i}(x^{*})\geq 0 \text{ for } \forall i \in \{1,..,k\}$
\item $\lambda_{i}c_{i}(x^{*})=0 \text{ for } \forall i \in \{1,..,k\}$
\end{enumerate}
$\lambda_{1},..,\lambda_{k}$ are known as Lagrange multipliers of constraints $c_{1},..,c_{k}$.
\end{prop}

The following well-known result states that KKT conditions are sufficient for 
$x^{*}$ to be a point with no feasible descent directions:
\begin{prop}
\label{prop-feas-desc}
If $\exists \lambda_{1},..,\lambda_{k}$ such that the following conditions are satisfied at $x^{*}$:
\begin{enumerate}
\item $\nabla f(x^{*})=\sum_{i=1}^{k}\lambda_{i}\nabla c_{i}(x^{*})$
\item $\lambda_{i}\geq 0 \text{ for } \forall i \in \{1,..,k\}$
\end{enumerate}
then $x^{*}$ has no feasible descent directions in the problem (\ref{eqn_constr_opt}) 
%($s$ is a feasible direction means that $\exists \epsilon>0:\text{ for }\forall \nu \in (0,\epsilon),\nu s$ is feasible?).

\begin{proof}\emph{(sketch)}
We reproduce the proof given in a textbook by Fletcher \citeyear{Fletcher}. The proposition is true because for any feasible direction vector $s$, $s^{T}\nabla c_{i}(x)\geq 0 \text{ for }\forall x \text{ and for }\forall i\in\{1,..,k\}$.  Hence, $s^{T}\nabla f(x^{*})=\sum_{i=1}^{k}\lambda_{i}s^{T}\nabla c_{i}(x^{*})\geq 0$, so $s$ is not a descent direction.
\end{proof}
\end{prop}

The following lemma characterizes the points in the set $\Gamma$:
\begin{lem}
\label{lem-gamma}
Let $\mu^{*} \in\Gamma$, and let $[w^{*},b^{*}]=H_{1}(\mu^{*})$ be the optimizer of $\psi(\mu^{*})$, and let $\boldsymbol{\lambda}^{*}=[\lambda^{*}_{\text{(\ref{eqn_constr_a1})},1},..,\lambda^{*}_{\text{(\ref{eqn_constr_a1})},m}, \lambda_{\text{(\ref{eqn_constr_a2})}}^{*},\lambda_{\text{(\ref{eqn_constr_a3})}}^{*}]$ be a set of Lagrange multipliers corresponding to the constraints for the solution $[w^{*},b^{*}]$. 
%($\boldsymbol{\lambda}^{(t)}$ may not be unique).  
Define $\mu'=H(\mu^{*})$, and let $[w',b']$ be the optimizer of $\psi(\mu')$.  
If $\mu'_{2} \neq \mu^{*}_{2}$, then $\lambda_{\text{(\ref{eqn_constr_a3})}}^{*}=0$ for some $\boldsymbol{\lambda}^{*}$. If $\mu'_{1} \neq \mu_{1}^{*}$, then $\lambda_{\text{(\ref{eqn_constr_a2})}}^{*}=0$ for some $\boldsymbol{\lambda}^{*}$.  
If both $\mu'_{1}\neq \mu_{1}^{*}\text{ and }\mu'_{2} \neq \mu_{2}^{*}$, then $\lambda_{\text{(\ref{eqn_constr_a3})}}^{*}=\lambda_{\text{(\ref{eqn_constr_a2})}}^{*}=0$ for some $\boldsymbol{\lambda}^{*}$.  
\begin{proof}
Consider the case when 
\begin{equation}
\mu'_{2}\neq \mu_{2}^{*}
\label{eqn_case1_1}
\end{equation}
and
\begin{equation}
\mu'_{1}=\mu_{1}^{*}
\label{eqn_case1_2}
\end{equation}
Since $\mu^{*}\in\Gamma$, $\left\Vert w' \right\Vert=\left\Vert w^{*} \right\Vert$.
Let $\boldsymbol{\lambda}'$ be a set of Lagrange multipliers corresponding to the constraints for the solution $[w',b']$. 
Since $w^{*}$ is still feasible for the optimization problem given by $\psi(\mu')$ (by the argument in Theorem \ref{thm-obj-conv}) and the minimum of this problem is unique, this can only happen if 
\begin{equation}
[w',b']=[w^{*},b^{*}].
\label{eqn_equal_w}
\end{equation}
Then $[w^{*},b^{*}]$ and $\boldsymbol{\lambda}'$ must satisfy KKT conditions for $\psi(\mu')$. (\ref{eqn_case1_1}) implies that \\$c_{1}(w^{*};\mu'_{2},\Sigma_{2})>c_{1}(w^{*};\mu_{2}^{*},\Sigma_{2})\geq \beta$ by the same argument as in Theorem \ref{thm-obj-conv}, which means that, by KKT condition (4) for $\psi(\mu')$, 
\begin{equation}
\lambda'_{\text{(\ref{eqn_constr_a3})}}=0.
\label{eqn_zero_lambda}
\end{equation}
Therefore, by KKT condition (1) for $\psi(\mu')$ and (\ref{eqn_zero_lambda}), at $[w,b,\mu_{1},\mu_{2}]=[w^{*}=w',b^{*}=b',\mu_{1}^{*}=\mu'_{1},\mu_{2}^{*}]$
\begin{align*}
\left[
\begin{array}{c}
\pd{\left\Vert w \right\Vert}{w}\\
\pd{\left\Vert w \right\Vert}{b}\\
\end{array}
\right]=
%\left[
%\begin{array}{c}
%\pd{\left\Vert w \right\Vert}{w}\\
%0\\
%\end{array}
%\right]=
\sum_{i=1}^{m}\lambda'_{\text{(\ref{eqn_constr_a1})},i}
\left[
\begin{array}{c}
y_{i}x_{i}\\
y_{i}\\
\end{array}\right]+
\lambda'_{\text{(\ref{eqn_constr_a2})}}\left[
\begin{array}{c}
\pd{c_{-1}(w,b^{*};\mu_{1}^{*},\Sigma_{1})}{w}\\
\pd{c_{-1}(w^{*},b;\mu_{1}^{*},\Sigma_{1})}{b}\\
\end{array}
\right]+
\lambda'_{\text{(\ref{eqn_constr_a3})}}\left[
\begin{array}{c}
\pd{c_{1}(w,b^{*};\mu_{2}^{*},\Sigma_{2})}{w}\\
\pd{c_{1}(w^{*},b;\mu_{2}^{*},\Sigma_{2})}{b}\\
\end{array}
\right],
\end{align*}
which means that KKT conditions (1),(2) for the optimization problem $\psi(\mu^{*})$ are satisfied at the point $[w^{*},b^{*}]$
with $\boldsymbol{\lambda}^{*}=\boldsymbol{\lambda}^{'}$.  KKT condition (3) is satisfied by feasibility of $[w^{*},b^{*}]$ and KKT condition (4) is satisfied by the same condition for $\psi(\mu')$ and observations (\ref{eqn_case1_2}), (\ref{eqn_equal_w}), and  (\ref{eqn_zero_lambda}).

The proofs for the other two cases ($\mu_{2}'=\mu_{2}^{*},\mu_{1}'\neq \mu_{1}^{*}$ and $\mu_{2}'\neq\mu_{2}^{*},\mu_{1}'\neq \mu_{1}^{*}$) are analogous.
\end{proof}
\end{lem}

The following theorem states that the points in $\Gamma$ are KKT points (i.e., points at which KKT conditions are satisfied)
for the optimization problem given by (\ref{eqn_orig_opt}).
\begin{thm}
\label{thm-kkt-points}
If $\mu^{*}\in\Gamma$ and let $[w^{*},b^{*}]=H_{1}(\mu^{*})$, then $[w^{*},b^{*},\mu_{1}^{*},\mu_{2}^{*}]$ is a KKT point for the optimization problem given by (\ref{eqn_orig_opt}).
\begin{proof}
Let $\mu'=H(\mu^{*})$.
Just like in Lemma \ref{lem-gamma}, we only consider the case 
\begin {equation}
\mu'_{2}\neq \mu_{2}^{*}, \label{eqn_ass1}
\end{equation}
\begin {equation}
\mu'_{1}=\mu_{1}^{*} \Rightarrow \lambda^{*}_{\text{(\ref{eqn_constr_a3})}}=0
\text{ (by Lemma \ref{lem-gamma})}.  \label{eqn_ass2}
\end{equation}
(the proofs for the other two cases are similar).

By KKT conditions for $H_{2}(w^{*},b^{*})$, at $\mu_{1}=\mu'_{1}$
%(\ref{eqn_alg_map2})-(\ref{eqn_constr_b2}) and (\ref{eqn_ass2}), 
%$\pd{(-(c_{-1}(w^{(t)},b^{(t)};\mu'_{1},\Sigma_{1})+c_{1}(w,b;\mu'_{2},\Sigma_{2})))}{\mu_{1}}=$
\begin{equation}
-\pd{c_{-1}(w^{*},b^{*};\mu_{1},\Sigma_{1})}{\mu_{1}}=\lambda'_{\text{\ref{eqn_constr_b1}}}\pd{(-o(\mu_{1};\Omega_{1},t))}{\mu_{1}}\text{ for some }
\lambda'_{\text{\ref{eqn_constr_b1}}}\geq0.
\label{eqn_kkt1}
\end{equation}

By KKT conditions for $H_{1}(\mu^{*})$ and (\ref{eqn_ass2}), at $[w,b]=[w^{*},b^{*}]$
%(\ref{eqn_alg_map1})-(\ref{eqn_constr_a4}),
\begin{align}
\left[
\begin{array}{c}
\pd{\left\Vert w \right\Vert}{w}\\
\pd{\left\Vert w \right\Vert}{b}\\
\end{array}
\right]=
\sum_{i=1}^{m}\lambda^{*}_{\text{(\ref{eqn_constr_a1})},i}
\left[
\begin{array}{c}
y_{i}x_{i}\\
y_{i}
\end{array}
\right]+
\lambda^{*}_{\text{(\ref{eqn_constr_a2})}}
\left[
\begin{array}{c}
\pd{c_{-1}(w,b^{*};\mu_{1}^{*},\Sigma_{1})}{w}\\
\pd{c_{-1}(w^{*},b;\mu_{1}^{*},\Sigma_{1})}{b}\\
\end{array}
\right]
\text{ for some }
\left[
\begin{array}{c}
\lambda^{*}_{\text{(\ref{eqn_constr_a1})},1}\\
..\\
\lambda^{*}_{\text{(\ref{eqn_constr_a1})},m}\\
\lambda^{*}_{\text{(\ref{eqn_constr_a2})}}
\end{array}
\right]\succeq 0.
\label{eqn_kkt2}
\end{align}

By (\ref{eqn_ass1}),(\ref{eqn_ass2}),(\ref{eqn_kkt1}), and (\ref{eqn_kkt2}),
at $[w,b,\mu_{1},\mu_{2}]=[w^{*},b^{*},\mu_{1}^{*}=\mu'_{1},\mu_{2}^{*}]$
\begin{align*}
\left[
\begin{array}{c}
\pd{\left\Vert w \right\Vert}{w}\\
\pd{\left\Vert w \right\Vert}{b}\\
\pd{\left\Vert w \right\Vert}{\mu_{1}}\\
\pd{\left\Vert w \right\Vert}{\mu_{2}}
\end{array}
\right]=
\left[
\begin{array}{c}
\pd{\left\Vert w \right\Vert}{w}\\
0\\
0\\
0
\end{array}
\right]=
\sum_{i=1}^{m}\lambda^{*}_{\text{(\ref{eqn_constr_a1})},i}
\left[
\begin{array}{c}
y_{i}x_{i}\\
y_{i}\\
0\\
0\end{array}\right]+
\lambda^{*}_{\text{(\ref{eqn_constr_a2})}}\left[
\begin{array}{c}
\pd{c_{-1}(w,b^{*};\mu_{1}^{*},\Sigma_{1})}{w}\\
\pd{c_{-1}(w^{*},b;\mu_{1}^{*},\Sigma_{1})}{b}\\
\pd{c_{-1}(w^{*},b^{*};\mu_{1},\Sigma_{1})}{\mu_{1}}\\
%\pd{(-o(\mu'_{1};\Omega_{1},t))}{\mu_{1}}\\
0
\end{array}
\right]+\\
\lambda'_{\text{\ref{eqn_constr_b1}}}\lambda^{*}_{\text{(\ref{eqn_constr_a2})}}
\left[
\begin{array}{c}
0\\
0\\
\pd{(-o(\mu_{1};\Omega_{1},t))}{\mu_{1}}\\
0\\
\end{array}
\right]+
\lambda^{*}_{\text{(\ref{eqn_constr_a3})}}\left[
\begin{array}{c}
\pd{c_{1}(w,b^{*};\mu_{2}^{*},\Sigma_{2})}{w}\\
\pd{c_{1}(w^{*},b;\mu_{2}^{*},\Sigma_{2})}{b}\\
0\\
\pd{c_{1}(w^{*},b^{*};\mu_{2},\Sigma_{2})}{\mu_{2}}\\
\end{array}
\right]+
\lambda^{*}_{\text{(\ref{eqn_constr_a3})}}
\left[
\begin{array}{c}
0\\
0\\
0\\
\pd{(-o(\mu_{2};\Omega_{2},t))}{\mu_{2}}\\
\end{array}
\right],
%\text{ which means that KKT conditions }
\end{align*}
which means that KKT conditions (1),(2) for the optimization problem (\ref{eqn_orig_opt}) are satisfied at the point $[w^{*},b^{*},\mu_{1}^{*},\mu_{2}^{*}]$
with $\boldsymbol{\lambda}^{''}=[
\lambda^{*}_{\text{(\ref{eqn_constr_a1})},1},..,\lambda^{*}_{\text{(\ref{eqn_constr_a1})},m},
\lambda^{*}_{\text{(\ref{eqn_constr_a2})}},
\lambda^{*}_{\text{(\ref{eqn_constr_a3})}},
\lambda'_{\text{\ref{eqn_constr_b1}}}\lambda^{*}_{\text{(\ref{eqn_constr_a2})}},
\lambda^{*}_{\text{(\ref{eqn_constr_a3})}}
]$.
$\boldsymbol{\lambda}^{''}$ also satisfies KKT conditions (3),(4) by assumption (\ref{eqn_ass2}) and the KKT conditions for $H_{1}$ and $H_{2}$.
\end{proof}
\end{thm}

In order to prove convergence properties of the iterates $\mu^{(t)},$
we use the following theorem due to Zangwill \citeyear{zangwill}:
%, which follows from the results of Meyer \citeyear{} and Ostrowsky \citeyear{}.
\begin{thm}
\label{thm-zangwill}
Let the map $H:Z\rightarrow Z$ determine an iterative algorithm via $\mu^{(t+1)}= H(\mu^{(t)})$, let $\psi(\mu)$ denote the objective function, and let $\Gamma$ be the set of points on which the map $H$ does not change the value of the objective function, i.e. $\mu \in \Gamma \Leftrightarrow \psi(H(\mu))=\psi(\mu)$.
Suppose 
\begin{enumerate}
\item $H$ is uniformly compact on $Z$, i.e. there is a compact subset $Z_{0}\subseteq Z$ such that $H(\mu) \in Z_{0}$ for $\forall \mu\in Z$.
\item $H$ is strictly monotonic on $Z-\Gamma$, i.e. $\psi(H(\mu))<\psi(\mu)$.
\item $H$ is closed on $Z-\Gamma$, i.e. if $w_{i} \rightarrow w$ and $H(w_{i}) \rightarrow \xi$, then $\xi=H(w)$.
\end{enumerate}
%Then either $\left\{z^{(t)}\right\}$ converges to a fixed point in $\Gamma$, or it has a continuum of accumulation points in $\Gamma$ (all with the same value of $\psi(\cdot)$).
Then the accumulation points of the sequence of $\mu^{(t)}$ lie in $\Gamma$.
\end{thm}

The following proposition shows that minimization of a continuous function on a feasible set which is a continuous map of the function's argument forms a closed function.
\begin{prop} 
\label{prop-closed}
Given 
\begin{enumerate}
\item a real-valued continuous function $f$ on $A\times B$,
\item a point-to-set map $U:A\rightarrow 2^{B}$ continuous with respect to 
the Hausdorff metric:\footnote{A point-to-set map $U(a)$ maps a point $a$ to a set of points.  $U(a)$ 
is continuous with respect to a distance metric $dist$ iff 
$a^{(t)}\rightarrow a$ implies $dist(U(a^{(t)}),U(a))\rightarrow 0$.}
$dist(X,Y)\triangleq \max(d(X,Y),d(Y,X)),$ 
where $d(X,Y)\triangleq \max_{x\in X}\min_{y \in Y}\left\Vert x-y \right\Vert$,
\end{enumerate}
 define the function $F:A\rightarrow B$ by 
\[F(a)=\arg\min_{b' \in U(a)}f(a,b')=\{b: f(a,b) < f(a,b') \text{ for }\forall b' \in U(a)\},\] assuming the minimum exists and is unique.  Then, the function $F$ is closed at $a$.
\begin{proof}
This proof is a minor modification of the one given by 
Gunawardana and Byrne \citeyear{gem}.  
Let $\{a^{(t)}\}$ be a sequence in $A$  such that 
\begin {equation}a^{(t)}\rightarrow a,F(a^{(t)})\rightarrow b
\label{eqn_conv}
\end{equation} 
The function $F$ is closed at $a$ if $F(a)=b$.
Suppose this is not the case, i.e. $b \neq F(a)=\arg\min_{b'\in U(a)}f(a,b')$.  Therefore, 
\begin{equation}
\exists \hat{b} =\arg\min_{b' \in U(a)}f(b') \text{ such that } f(a,b)>f(a,\hat{b})
\label{eqn_min}
\end{equation}
By continuity of $f(\cdot,\cdot)$ and (\ref{eqn_conv}),
\begin{equation}
f(a^{(t)},F(a^{(t)}))\rightarrow f(a,b)
\label{eqn_e1}
\end{equation}
By continuity of $U(\cdot)$ and (\ref{eqn_conv}),
\begin{equation}
dist(U(a^{(t)}),U(a))\rightarrow 0 \Rightarrow \exists \hat{b}^{(t)}\rightarrow \hat{b}
\text{ and }\hat{b}^{(t)}\in U(a^{t}), \text{ for }\forall t.
\label{eqn_e2}
\end{equation}
(\ref{eqn_e1}), (\ref{eqn_e2}), and (\ref{eqn_min}) imply that 
\begin{equation}
\exists K \text{ such that }
f(a^{(t)},F(a^{(t)}))>f(a^{(t)},\hat{b}^{(t)}),\text{ for }\forall t>K
\label{eqn_conv_hat}
\end{equation}
which is a contradiction since by assumption, $F(a^{(t)})=\arg\min_{b' \in U(a^{t})}f(b')$ and by (\ref{eqn_conv_hat}), $\hat{b}^{(t)} \in U(a^{(t)})$.
\end{proof}
\end{prop}

\begin{prop}
\label{prop-closed-q}
The function $H$ defined by (\ref{eqn_alg_map})-(\ref{eqn_alg_map2}) is closed.
\begin{proof}
Let $\{\mu^{(t)}\}$ be a sequence such that $\mu^{(t)}\rightarrow \mu^{*}$. 
Since all the iterates $\mu^{(t)}$ lie in the closed feasible region bounded by constraints (\ref{eqn1_constr1})-(\ref{eqn1_constr4}) and the boundary of $U(\mu)$ is piecewise linear in $\mu$, the boundary of $U(\mu^{(t)})$ converges uniformly to the boundary of $U(\mu^{*})$ as $\mu^{(t)}\rightarrow \mu^{*}$, which implies that the Hausdorff distance between the boundaries converges to zero.  Since the Hausdorff distance between convex sets is equal to the Hausdorff distance between their boundaries, $dist(U(\mu^{(t)}),U(\mu^{*}))$ also converges to zero.  Hence, proposition \ref{prop-closed} implies that 
$H_{1}$ is closed.  The same proposition implies that $H_{2}$ is closed.  A composition of closed functions is closed, hence $H$ is closed.
\end{proof}
\end{prop}
We now prove the main result of this Section:
\begin{thmb}
Let $H$ be the function defined by (\ref{eqn_alg_map})-(\ref{eqn_alg_map2}) which determines the generative/discriminative algorithm via $\mu^{(t+1)} =H (\mu^{(t)})$.  Then accumulation points $\mu^{*}$ of the sequence $\mu^{(t)}$ augmented with $[w^{*},b^{*}]=H_{1}(\mu^{*})$ have no feasible descent directions for the original optimization problem given by (\ref{eqn_opt})-(\ref{eqn1_constr4}).
\begin{proof}
The proof is by verifying that $H$ satisfies the properties of Theorem \ref{thm-zangwill}.  Closedness of $H$ was shown in Proposition \ref{prop-closed-q}.  Strict monotonicity of $\psi(\mu^{(t)})$ was shown in Theorem \ref{thm-obj-conv}.  Since all the iterates $\mu^{(t)}$ are in the closed feasible region bounded by constraints (\ref{eqn1_constr1})-(\ref{eqn1_constr4}), $H$ is uniformly compact on $Z$.  Since all the accumulation points $\mu^{*}$ lie in $\Gamma$, they are KKT points of the original optimization problem by Theorem \ref{thm-kkt-points}, and, therefore, have no feasible descent directions by Proposition \ref{prop-feas-desc}.
\end{proof}
\end{thmb}

\section{Generalization of the Generative/Discriminative Classifier}
\label{appendix-gen-bound}
We need a few auxiliary results before proving Theorem \ref{thm-gen-bound}.
The first proposition bounds the angle of rotation
between two vectors $w_{1},w_{2}$ and the distance between them if
the angle of rotation between each of these vectors and some reference
vector $v$ is sufficiently small:

\begin{prop}
\label{prop-angle-bound}
Let $\left\Vert w_{1}\right\Vert =\left\Vert w_{2}\right\Vert =\left\Vert v\right\Vert =1$.
If $w_{1}^{T}v\geq\alpha\geq0$ and $w_{2}^{T}v\geq\alpha\geq0$, then 
\begin{enumerate}
\item $w_{1}^{T}w_{2}\geq2\alpha^{2}-1$ 
\item $\left\Vert w_{1}-w_{2}\right\Vert \leq2\sqrt{(1-\alpha^{2})}$
\end{enumerate}
\begin{proof}
~

\begin{enumerate}
\item By the triangle inequality, $\arccos(w_{1}^{T}w_{2})\leq\arccos(w_{1}^{T}v)+\arccos(w_{2}^{T}v)\leq2\arccos(\alpha)$
(since the angle between two vectors is a distance measure). Taking
cosines of both sides and using trigonometric equalities yields $w_{1}^{T}w_{2}\geq2\alpha^{2}-1$.
\item Expand $\left\Vert w_{1}-w_{2}\right\Vert ^{2}=\left\Vert w_{1}\right\Vert ^{2}+\left\Vert w_{2}\right\Vert ^{2}-2w_{1}^{T}w_{2}=2(1-w_{1}^{T}w_{2})$.
Since $w_{1}^{T}w_{2}\geq2\alpha^{2}-1$ from part 1, $\left\Vert w_{1}-w_{2}\right\Vert ^{2}\leq4(1-\alpha^{2})$.
\end{enumerate}
\end{proof}
\end{prop}

The next proposition bounds the angle of rotation between two vectors
$t$ and $\mu$ if they are not too far away from each other as measured
by the $L_{2}$-norm distance:

\begin{prop}
\label{prop-angle-bound2}
Let $\left\Vert t\right\Vert =\nu$, $\left\Vert \mu-t\right\Vert \leq\tau$.
Then $\frac{t^{T}\mu}{\left\Vert t\right\Vert \left\Vert \mu\right\Vert }\geq\frac{\nu^{2}-\tau^{2}}{\nu(\nu+\tau)}$.
\begin{proof}
Expanding $\left\Vert \mu-t\right\Vert ^{2}=\left\Vert t\right\Vert ^{2}+\left\Vert \mu\right\Vert ^{2}-2t^{t}\mu$
and using $\left\Vert \mu-t\right\Vert ^{2}\leq\tau^{2}$, we get $\frac{t^{T}\mu}{\left\Vert t\right\Vert \left\Vert \mu\right\Vert }\geq\frac{1}{2}(\frac{\left\Vert t\right\Vert }{\left\Vert \mu\right\Vert }+\frac{\left\Vert \mu\right\Vert }{\left\Vert t\right\Vert }-\frac{\tau^{2}}{\left\Vert t\right\Vert \left\Vert \mu\right\Vert })$.
We now use the triangle inequality $\nu-\tau\leq\left\Vert t\right\Vert -\left\Vert \mu-t\right\Vert \leq\left\Vert \mu\right\Vert \leq\left\Vert t\right\Vert +\left\Vert \mu-t\right\Vert \leq\nu+\tau$
and simplify.
\end{proof}
\end{prop}

The following proposition will be used to bound the angle of rotation
between the normal $w$ of the separating hyperplane and the mean
vector $t$ of the hyper-prior distribution:

\begin{prop}
\label{prop-hyperprior-angle}
Let $\frac{w^{T}\mu}{\left\Vert w\right\Vert \left\Vert \mu\right\Vert }\geq\beta\geq0$
and $\left\Vert \mu-t\right\Vert \leq\varphi\leq\left\Vert t\right\Vert $.
Then $\frac{w^{T}t}{\left\Vert w\right\Vert \left\Vert t\right\Vert }\geq(2\alpha^{2}-1)$,
where $\alpha=\min(\beta,\frac{\left\Vert t\right\Vert ^{2}-\varphi^{2}}{\left\Vert t\right\Vert (\varphi+\left\Vert t\right\Vert )})$.
\begin{proof}
Follows directly from Propositions \ref{prop-angle-bound} (part 1) and \ref{prop-angle-bound2}.
\end{proof}
\end{prop}

We now prove Theorem \ref{thm-gen-bound}, which relies on parts of the well-known proof of the fat-shattering dimension bound for large margin classifiers derived by Taylor and Bartlett \citeyear{bartlett:generalize}.

\begin{thma}
Let $F$ be the class of a-priori constrained functions defined by \ref{eqnf},
and let $\lambda_{min}(P)$ and $\lambda_{max}(P)$ denote the minimum and maximum eigenvalues of matrix $P$, respectively.
If a set of points $S$ is $\gamma$-shattered by $F$, then
$\left|S\right|\leq\frac{4R^{2}(\alpha^{2}(1-\alpha^{2}))}{\gamma^{2}}$,
where $\alpha=\max(\alpha_{1},\alpha_{2})$ with $\alpha_{1}=\min(\frac{\lambda_{min}(\Sigma_{1})\beta}{\left\Vert \mu_{2}\right\Vert },\frac{\left\Vert t_{2}\right\Vert ^{2}-(\lambda_{max}(\Omega_{2})\varphi)^{2}}{\left\Vert t_{2}\right\Vert (\lambda_{max}(\Omega_{2})\varphi)^{2}+\left\Vert t_{2}\right\Vert )})$
and $\alpha_{2}=\min(\frac{\lambda_{min}(\Sigma_{1})\beta}{\left\Vert \mu_{1}\right\Vert },$
$\frac{\left\Vert t_{1}\right\Vert ^{2}-(\lambda_{max}(\Omega_{1})\varphi)^{2}}{\left\Vert t_{1}\right\Vert ((\lambda_{max}(\Omega_{1})\varphi)^{2}+\left\Vert t_{1}\right\Vert )})$,
assuming that $\beta\geq0$, $\left\Vert t_{i}\right\Vert \geq\left\Vert t_{i}-\mu_{i}\right\Vert $,
and $\alpha_{i}\geq\frac{1}{\sqrt{2}},i=1,2$. 
\begin{proof}
First, we use the inequality $\lambda_{min}(P)\left\Vert w\right\Vert \leq\left\Vert P^{1/2}w\right\Vert \leq \lambda_{max}(P)\left\Vert w \right\Vert$ 
%\leq\lambda_{max}(P)\left\Vert w\right\Vert $,
%where $\lambda_{max}(P)$ and $\lambda_{min}(P)$ are the maximum
%where $\lambda_{min}(P)$ is the minimum eigenvalue of $P$ 
to relax the constraints 
\begin{equation}
\label{eqn4}
\frac{w^{T}\mu_{2}}{\left\Vert \Sigma_{2}^{1/2}w\right\Vert }\geq\beta\Rightarrow\frac{w^{T}\mu_{2}}{\left\Vert w\right\Vert }\geq\lambda_{min}(\Sigma_{2})\beta
\end{equation}
\begin{equation}
\label{eqn5}
\left\Vert \Omega_{2}^{-1/2}(\mu_{2}-t_{2})\right\Vert \leq\varphi\Rightarrow\left\Vert \mu_{2}-t_{2}\right\Vert \leq\frac{\varphi}{\lambda_{min}(\Omega_{2}^{-1})}=\varphi\lambda_{max}(\Omega_{2}).
\end{equation}
The constraints imposed by the second prior $\frac{-w^{T}\mu_{1}}{\left\Vert \Sigma_{2}^{1/2}w\right\Vert }\geq\beta,\left\Vert \Omega_{1}^{-1/2}(\mu_{1}-t_{1})\right\Vert \leq\varphi$
are relaxed in a similar fashion to produce:
\begin{equation}
\label{eqn41}
\frac{w^{T}(-\mu_{1})}{\left\Vert w\right\Vert }\geq\lambda_{min}(\Sigma_{1})\beta
\end{equation}
\begin{equation}
\label{eqn51}
\left\Vert \mu_{1}-t_{1}\right\Vert \leq\varphi\lambda_{max}(\Omega_{1})
\end{equation}
Now, we show that if the assumptions made in the statement of the
theorem hold, then every subset $S_{o}\subseteq S$ satisfies $\left\Vert \sum S_{o}-\sum(S-S_{0})\right\Vert \leq\frac{4R^{2}(\alpha^{2}(1-\alpha^{2})}{\gamma^{2}}$.

Assume that $S$ is $\gamma$-shattered by $F$. The argument used
by Taylor and Bartlett \citeyear{bartlett:generalize} in
Lemma 1.2 shows that, by the definition
of fat-shattering, there exists a vector $w_{1}$ such that 
\begin{equation}
w_{1}(\sum S_{o}-\sum(S-S_{0}))\geq\left|S\right|\gamma.
\end{equation}
Similarly (reversing the labeling of $S_{0}$ and $S_{1}-S_{0}$),
there exists a vector $w_{2}$ such that 
\begin{equation}
w_{2}(\sum(S-S_{0})-\sum S_{o})\geq\left|S\right|\gamma.
\end{equation}
Hence, $(w_{1}-w_{2})(\sum S_{o}-\sum(S-S_{0}))\geq2\left|S\right|\gamma$,
which, by Cauchy-Schwartz inequality, implies that 
\begin{equation}
\label{eqn6}
\left\Vert w_{1}-w_{2}\right\Vert \geq\frac{2\left|S\right|\gamma}{\left\Vert \sum S_{o}-\sum(S-S_{0})\right\Vert }
\end{equation}
The constraints on the classifier represented in \ref{eqn4} and \ref{eqn5}
imply by Proposition \ref{prop-hyperprior-angle} that $\frac{w_{1}^{T}t_{2}}{\left\Vert w_{1}\right\Vert \left\Vert t_{2}\right\Vert }\geq(2\alpha_{1}^{2}-1)$
and $\frac{w_{2}^{T}t_{2}}{\left\Vert w_{2}\right\Vert \left\Vert t_{2}\right\Vert }\geq(2\alpha_{2}^{2}-1)$
. Now, applying Proposition \ref{prop-angle-bound} (part 2) and simplifying, we get 
\begin{equation}
\label{eqn7}
\left\Vert w_{1}-w_{2}\right\Vert \leq4\sqrt{\alpha_{1}^{2}(1-\alpha_{1}^{2})}.
\end{equation}
Applying the same analysis to the constraints \ref{eqn41} and \ref{eqn51},
we get 
\begin{equation}
\label{eqn8}
\left\Vert w_{1}-w_{2}\right\Vert \leq4\sqrt{\alpha_{2}^{2}(1-\alpha_{2}^{2})}.
\end{equation}

Combining \ref{eqn6}, \ref{eqn7}, and \ref{eqn8}, we get 
\begin{equation}
\label{eqn9}
\left\Vert \sum S_{o}-\sum(S-S_{0})\right\Vert \geq\frac{\left|S\right|\gamma}{2\sqrt{\alpha^{2}(1-\alpha^{2})}}
\end{equation} with $\alpha$ as defined in the statement of the theorem.

Taylor and Bartlett's \citeyear{bartlett:generalize} Lemma 1.3 proves, using the probabilistic
method, that some $S_{o}\subseteq S$ satisfies 
\begin{equation}
\label{eqn10}
\left\Vert \sum S_{o}-\sum(S-S_{0})\right\Vert \leq\sqrt{\left|S\right|}R. 
\end{equation}
Combining \ref{eqn9} and \ref{eqn10} yields $\left|S\right|\leq\frac{4R^{2}(\alpha^{2}(1-\alpha^{2}))}{\gamma^{2}}$.
\end{proof}
%\tag{\ref{thm-gen-bound}}
\end{thma}
\bibliographystyle{theapa}
\bibliography{newfile1}
%\begin{thebibliography}{10}
%\bibitem[\protect\BCAY{Tong\ \BBA\ Koller}{Tong\ \BBA\ Koller}{2000b}]{svm}
%Tong, S.\BBACOMMA\  \BBA\ Koller, D. \BBOP2000b\BBCP.
%\newblock \BBOQ Support vector machine active learning with applications to
%  text classification\BBCQ\
%\newblock In {\Bem Proceedings of ICML}.
%\end{thebibliography}
\end{document}